\documentclass[final,1p,authoryear, 11pt]{elsarticle}

\usepackage{amssymb}
\usepackage[colorlinks,linkcolor=blue]{hyperref}
\usepackage{ragged2e}
\usepackage{bookmark}
\usepackage{indentfirst} 
\usepackage{geometry}
\usepackage{graphicx} 
\usepackage{float}
\usepackage{subfigure} 
\usepackage{amsmath}
\usepackage{booktabs}
\usepackage{diagbox}
\usepackage{tabularx}
\usepackage{amssymb}
\usepackage{makecell}
\usepackage{multirow}
\usepackage{mathptmx}
\usepackage{amsfonts}
\usepackage{chemformula}
\usepackage[T1]{fontenc}
\usepackage{caption}

\usepackage{boxedminipage}
\usepackage{textpos}

\journal{Computers, Environment and Urban Systems}

\begin{document}

\begin{frontmatter}

\title{Classification of Urban Morphology with Deep Learning: Application on Urban Vitality}

\author[doa]{Wangyang Chen}
\ead{e0403826@u.nus.edu}
\author[doa]{Abraham Noah Wu}
\ead{abraham@nus.edu.sg}
\author[doa,dre]{Filip Biljecki\corref{cor1}}
\ead{filip@nus.edu.sg}

\address[doa]{Department of Architecture, National University of Singapore, Singapore}
\address[dre]{Department of Real Estate, National University of Singapore, Singapore}

\cortext[cor1]{Corresponding author}

\begin{abstract}

There is a prevailing trend to study urban morphology quantitatively thanks to the growing accessibility to various forms of spatial big data, increasing computing power, and use cases benefiting from such information. 
The methods developed up to now measure urban morphology with numerical indices describing density, proportion, and mixture, but they do not directly represent morphological features from the human's visual and intuitive perspective.
We take the first step to bridge the gap by proposing a deep learning-based technique to automatically classify road networks into four classes on a visual basis.
The method is implemented by generating an image of the street network (Colored Road Hierarchy Diagram), which we introduce in this paper, and classifying it using a deep convolutional neural network (ResNet-34).
The model achieves an overall classification accuracy of 0.875.
Nine cities around the world are selected as the study areas with their road networks acquired from OpenStreetMap.
Latent subgroups among the cities are uncovered through clustering on the percentage of each road network category.
In the subsequent part of the paper, we focus on the usability of such classification: we apply our method in a case study of urban vitality prediction.
An advanced tree-based regression model (LightGBM) is for the first time designated to establish the relationship between morphological indices and vitality indicators.
The effect of road network classification is found to be small but positively associated with urban vitality.
This work expands the toolkit of quantitative urban morphology study with new techniques, supporting further studies in the future.
\end{abstract}




\end{frontmatter}


\begin{textblock*}{\textwidth}(0cm,-21.3cm)
\begin{center}
\begin{footnotesize}
\begin{boxedminipage}{1\textwidth}
This is the Accepted Manuscript version of an article published by Elsevier in the journal \emph{Computers, Environment and Urban Systems} in 2021, which is available at: \url{https://doi.org/10.1016/j.compenvurbsys.2021.101706}. Cite as:
Chen W, Wu AN, Biljecki F (2021): Classification of Urban Morphology with Deep Learning: Application on Urban Vitality. \textit{Computers, Environment and Urban Systems}, 90: 101706.
\end{boxedminipage}
\end{footnotesize}
\end{center}
\end{textblock*}

\begin{textblock*}{\textwidth}(-0.5cm,3.82cm)
{\tiny{\copyright{ }2021, Elsevier. Licensed under the Creative Commons Attribution-NonCommercial-NoDerivatives 4.0 International (\url{http://creativecommons.org/licenses/by-nc-nd/4.0/})}}
\end{textblock*}

\section{Introduction}
\label{Sec.1}
Urban morphology is the study of the formation of human settlements and the process of their formation and transformation \citep{moudon1997urban}.
It dissects the urban built environment into components (building, street, block, etc.) and aims to understand the running pattern among each component, the interaction between components, and influence on a multitude of phenomena such as urban microclimate and energy \citep{Yuan.2019,Garbasevschi.2021}.

The mainstream of quantitative morphological study operates by transforming urban entities into numerical indices.
There exists a wealth of literature concentrating on the quantification of the characters of building and space.
\citet{lynch1984good} quantitatively studies the relationship among floor space, distribution of open space, building height, and other elements.
\citet{alexander1977pattern} lists and explains the indicators of urban morphology at the city, neighborhood, and building levels but has not provided the corresponding calculation methods.
\citet{berghauser2007spacemate} propose a graph-based approach to analyze the urban form on the ground of four major morphological indicators and denominated this method as ``spatial matrix'' \citep{berghauser2010spacematrix}.
Other scholars take road networks as the point of departure.
Network-based methods represented by spatial syntax are put forward to analyze the configuration of networks \citep{hillier1984social,hillier1996space}.
The analytical techniques of space syntax have been utilized in various studies related to urban morphology \citep{jiang2002integration,baran2008space,ye2014quantitative}.
Further, many studies attempt to set up their morphological index system \citep{bocher2018geoprocessing,li2020urban,li2020data}, especially the recent study of \citet{martino2021urban}, which includes nearly every measurement mentioned above.
While these methods have contributed to explaining urban morphology in a more scientific and mathematical manner, they usually rely on calculations on vector spatial data.

The advent of new techniques such as deep learning, which are gaining momentum in urban studies \citep{Wu.2020pdi,Zhang.2020,Ding.2021}, may enable more innovative ways to measure urban morphology.
In this paper, we posit that these techniques are still underexplored in the domain of characterizing the urban form.
Therefore, we aim to develop an approach to extract features from morphological images with convolutional neural networks, demonstrating the usability of new technologies such as deep learning in the perennial topic of urban morphology.

As a case study, we focus on urban vitality, which has been a prevalent topic for decades.
The work of \citet{jacobs1961death} conceptualizes urban vitality as the capacity of the living environment to motivate the interactions among people and between people and the environment. We adopt this definition in this paper. The newly available open spatial big data sources (as opposed to traditional data collected from surveys) provide a great convenience for the study of urban vitality in the urban area, continuously maintaining it topical.
In fact, many instances of research related to the measurement of urban vitality have been conducted based on different kinds of data sources.

There are two major approaches to assess the intensity of urban vitality, from the perspective of the built environment and people, respectively.
One is by measuring the capacity of the built environment to boost activities \citep{lopes2013public, delclos2018looking, yue2019spatial,jin2017evaluating,ye2018block,zeng2018spatially}, while the other one is by measuring the density of people \citep{kim2020data,chen2019identifying,wu2018check,yang2021elaborating,meng2019exploring,yue2017measurements,li2020urban}. 
Meanwhile, researchers also adopted a hybrid method to carry out their assessment \citep{xia2020analyzing,he2018impact,botta2021modelling}.
According to the definition, urban vitality relates to two aspects, human and environment. As we cannot guarantee that an individual data source reflects each aspect sufficiently well, we need to involve multiple data sources for each of them.

Many previous studies have been devoted to delineating the relationship between urban morphology and vitality.
Regarding the model to establish the relationship, Ordinary Least Squares (OLS) regression is the most widespread model, as researchers have preferred its simplicity and interpretability \citep{sung2015residential,ye2018block,yue2017measurements,meng2019exploring,wu2018check,zarin2015physical}.
However, the drawback of OLS is also obvious.
On the one hand, OLS regression was noted to disregard the spatial dependence among the places which is deemed commonly existing \citep{xia2020analyzing,li2020urban}.
On the other hand, the linearity hypothesis on the relationship between urban morphology and urban vitality was challenged \citep{yang2021elaborating}.
To alleviate the problem, it might be worthwhile to apply a more advanced model to describe the relationship.

In this study, tackling the research opportunities and shortcomings described so far, we mainly aim to fill the following gaps and present the following contributions to the field:
(1) we introduce a supervised deep learning method to classify road networks from human perception, with the outlook of supporting a wide range of applications in urban morphology studies;
(2) we adopt the method in a case study of urban vitality prediction and investigate its effectiveness;
(3) we propose additional dimensions to indicate urban vitality and mitigate biases caused by single indicator, alleviating one of the key shortcomings of related work; and
(4) we adopt a more advanced model to delineate the relationship between urban morphology and urban vitality.

The first part of the work --- the development of an automated approach to characterize urban form using deep learning --- is independent, as it is devoted to establishing a universal method that may be applicable across many quantitative morphology studies in future.
The subsequent part, regarding urban vitality, is a case study picked out from the multitudinous possible applications of our method, and with the advancements we outline, we believe that it is a contribution on its own.

The remainder of this paper is organized as follows.
Section~\ref{Sec.2} reviews related research on the measurement of urban morphology and urban vitality, and the methods to uncover the relationship between them.
Section~\ref{Sec.3} introduces the framework, process of data curation, and corresponding models of our method.
Section~\ref{Sec.4} presents the research results including the performance of the models and the outcomes, followed by the interpretation of the results.
Section~\ref{Sec.5} discusses the results and describes limitations and future studies.
Section~\ref{Sec.6} concludes the paper.

\section{Background and related work}
\label{Sec.2}
\subsection{Urban Morphology}

Novel methods and tools were developed to support studying urban morphology \citep{Fleischmann:2019fr,Fleischmann:2020fe,Jochem.2021}.
For example, OSMnx provides an efficient way to acquire massive data of built environments from OpenStreetMap and analyze them \citep{BOEING2017126}.
\citet{boeing2021spatial} demonstrates a workflow of its application on street pattern, orientation and configuration studies.
With these tools, it is now possible to analyze street networks across thousands of urban areas \citep{boeing2020multi}.

Urban morphology characterized by morphological indices conveys quantities but overlooks the pattern that could be grasped by eyes.
The flourishing development of deep learning techniques enables machines to get a humanlike perception of urban morphology based on images.
In this field, related research efforts have been carried out in the microscale built environment.
They concentrated on object detection or segmentation of street view images to study street-level morphological features \citep{middel2019urban}, demographic make-up \citep{gebru2017using}, landscapes \citep{helbich2019using, kim2021decoding}, walkability \citep{zhou2019social}, and neighborhood vitality \citep{wang2020life}.

However, very limited relevant studies are conducted in mesoscale or macroscale urban form studies, and not many have focused on the orthogonal view instead of the street view.
For instance, a fully-connected neural network is applied to optimize urban morphological indicators for intensive and sustainable development of urban blocks \citep{qu2019investigating}.
However, urban morphology is still measured by traditional indicators and no information is extracted from the human perspective.
To the best of our knowledge, the research of \citet{moosavi2017urban} is the only one that adopts deep learning methods to obtain morphological features from images of road networks.
The work involves one million cities and utilizes Convolutional Auto Encoder (CAE) to transform images of road networks into urban vectors and further classifies the cities.
However, the work was not published in a peer-reviewed international scientific journal, and due to the usage of an unsupervised learning method, the model was not given any human perception.
Coupled with the problem of excessive scale, the study suffers from low interpretability of outcomes and disability of providing practical recommendations for urban morphology design.
All these drawbacks call for a more granular morphological study based on a supervised deep learning method that can interpret morphology with human sense, which we seek to accomplish in our research.

\subsection{Urban Vitality}

Studies over the past several years include numerous big data-based methods to measure urban vitality from the perspective of built environment, density of people, and so on.

Points of interest (POI) is the most frequently used data source that describe the built environment that supports human activities.
Density of POI is a common indicator of urban vitality, which is oftentimes supplemented by other indicators including density of road junctions \citep{jin2017evaluating}, housing prices and population change \citep{he2018impact}, and other social-economic statistics \citep{zeng2018spatially}.
Some research also reckons that urban vitality can be gauged via small catering businesses \citep{xia2020analyzing, ye2018block}.
They assume that the places where small catering businesses flourish tend to be more vibrant because small catering businesses relied heavily on large flow of active people. 

The density of people is in most cases measured by location-based services (LBS) data.
For example, social media check-ins are used to indicate the distribution of urban vitality
\citep{wu2018check, chen2019identifying}.
\citet{meng2019exploring} leverage place-based reviews as an alternative.
\citet{yang2021elaborating} use the Baidu Heat Index (the density of geographical locations of Baidu mobile application users) to represent urban vitality.
Further studies highlight the usage of mobile phone positioning data to portray urban vitality \citep{yue2017measurements,li2020urban}.
Besides, \citet{kim2018seoul} adopts bankcard transaction data as the proxy of economic vitality and evaluated virtual vitality by Wi-Fi access points.

Nighttime light (NTL) data provides an additional approach to capture vitality.
Such data is proved to have a correlation with economic activities \citep{zhang2011mapping,ma2014diverse}.
Consequently, it is leveraged to detect ``ghost cities'' (cities with low urban vitality) \citep{jin2017evaluating, zheng2017monitoring, ge2018ghost}, and delineate nighttime urban vitality in street block level \citep{jin2017evaluating,xia2020analyzing}.
Even though NTL data has been testified somehow correlative with nighttime urban vitality, some reports of its inconsistency with real human activities \citep{jin2017evaluating, zhao2019applications} necessitate the complement of other data sources.

Some scholars attempt to evaluate urban vitality in a bigger picture.
\citet{landry2000urbann} assumes that urban vitality is related to the economic, social, and physical environment. 
\citet{liu2010evaluation} build up a system that evaluates economic vitality, social vitality, cultural vitality, and environmental vitality, supplemented by innovation vitality \citep{martino2021urban}.
However, the data source they use often lacks granularity.
Conveniently, the advent of data sources such as WorldPop and Airbnb facilitates acquiring relevant data with higher granularity \citep{worldpop,Tatem.2017,Li:2019vv}.
In our study, overcoming the shortcomings of related work, we leverage these datasets as new proxies of urban vitality.

\subsection{Urban Morphology and Urban Vitality}

The OLS regression model is the most commonly used method to analyze the relationship between urban morphology and vitality.
However, it suffers from two major deficiencies: overlook of spatial relationship and excessive hypothesis of linearity.
Some effort has been made to rectify these issues.
Spatial referenced models such as Local Moran’s I statistics \citep{xia2020analyzing} and geographically weighted regression (GWR) \citep{li2020urban} have been adopted to alleviate the problem of ignoring the spatial relationship.
However, the usage of nonlinear model is still an underexplored domain, and another gap we seek to fill in this study.
Morphological design in the future calls for better evaluation of urban morphology's influence on vitality.
Nonlinear models qualified with higher complexity are confirmed to be more accurate in urban study tasks such as land use classification \citep{cao2019comparison} and transportation mode classification \citep{xiao2017identifying}. Among their model sets, XGBoost achieves the best performance.
LightGBM, which is also from the family of gradient boosting models, is a further improved version proposed by \citet{ke2017lightgbm}.
LightGBM achieves further improvement of training speed over XGBoost when inheriting most of its merits.
Besides accuracy, the identification of key morphological indices is also fundamental for the improvement of urban morphology design in new constructions.
As a tree-based model, LightGBM is an appropriate approach enabling such competences.
This study would for the first time apply LightGBM to explore the relationship between urban morphology and vitality.

\section{Methodology}
\label{Sec.3}
\subsection{Research Framework}
\subsubsection{Analytical framework}

\begin{figure*}[tbp]
\centering
\includegraphics[width=0.8\textwidth]{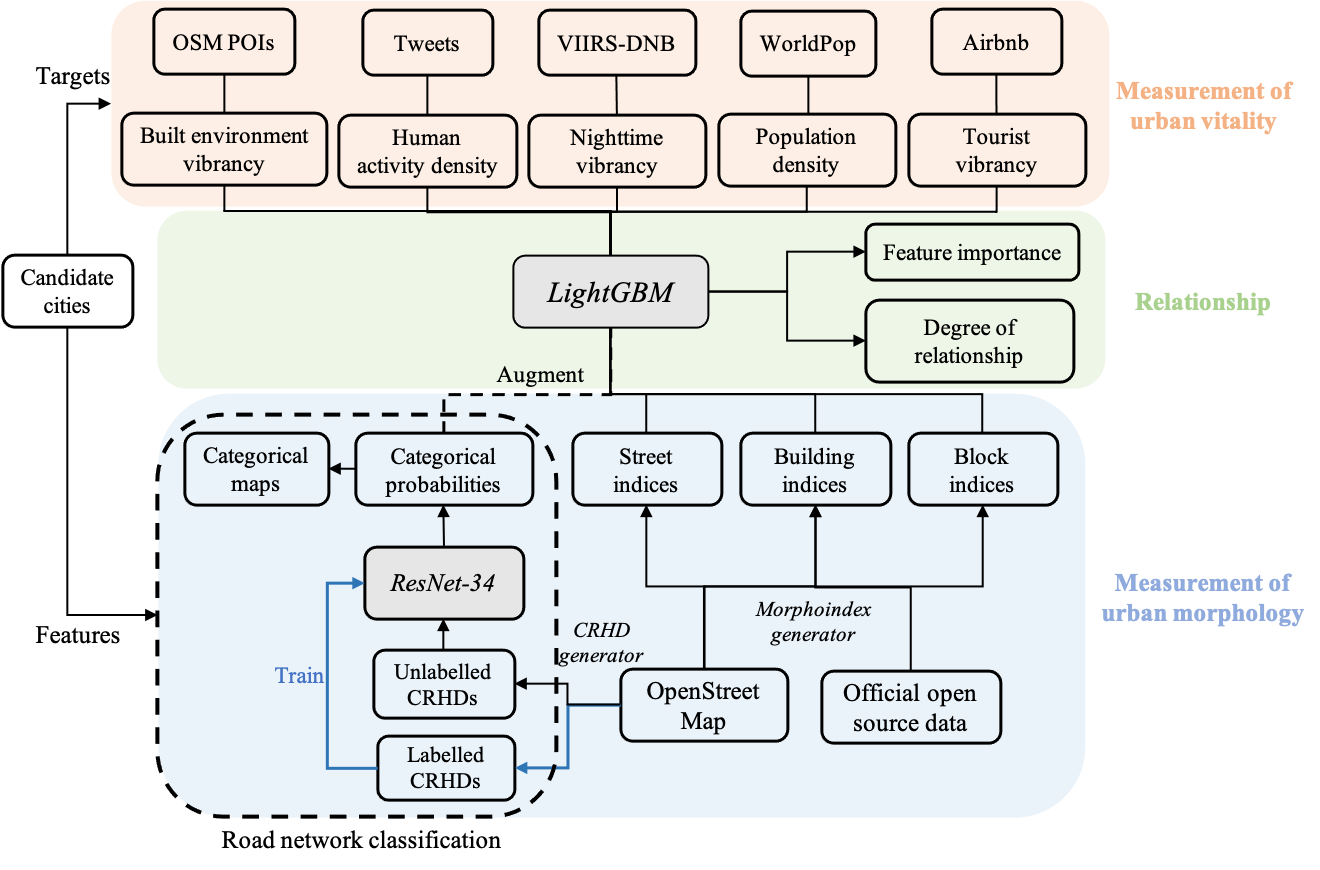}
\caption{Analytical framework of our work.}
\label{Fig.0}
\end{figure*}

Figure~\ref{Fig.0} illustrates the analytical framework of our work, which is composed of three parts: measurement of urban morphology, measurement of urban vitality, and the relationship between them.

The first step is intended to measure urban morphology through two approaches, a traditional one with morphological indices and a novel one we introduce, with road network classification (inside the dashed box).
Traditional morphological indices regarding street, building and block are calculated by the \textit{Morphoindex generator}, a Python script developed for automatic generation of morphological indices, which we implement as part of this study. The geospatial datasets used in this study stem from two main sources. One is OpenStreetMap (OSM), a Volunteered Geographic Information (VGI) platform with global coverage and of high quality in many areas around the world \citep{BarringtonLeigh:2017ie,Biljecki.2020}, and the others are the official open datasets. 

Our proposed classification is conducted in three steps. First, we generate the image of the road network legible for a computer. We develop a visual representation of the road network called Colored Road Hierarchy Diagram (CRHD) and implement its generation in Python.
The second step is to train the road network classification model, which is based on the ResNet-34 architecture.
A large group of CRHDs of representative urban zones are manually labelled and used for training.
After training, the model is able to calculate the probabilities of each road network category given an unlabeled CRHD.
With CRHDs classified based on the highest probability, we export categorical maps at the city scale.
The probabilities can be input into the feature space of quantitative morphology studies as an augmentation.
We maintain that it could inject simulated human perceptions that are oftentimes ignored in the field.

The effectiveness of the proposed method would be evaluated through a case study of urban vitality prediction.
Based on multi-source geospatial data, urban vitality is assessed from five main aspects: built environment vibrancy measured by POI density, human activity density measured by tweet density, nighttime vibrancy measured by nighttime light brightness, population density measured by population size, and tourism vibrancy measured by density of Airbnb listings.
This paper attaches equal importance to all the five aspects, and they are expected to reduce the latent biases for each other. We combine these indicators together as ``vitality score'' as a comprehensive evaluation of urban vitality.

The final portion of the paper aims to identify the correlation between urban morphology and urban vitality. 
This study adopts LightGBM, a tree-based gradient boosting method to delineate such relationships.
Vitality score is used to evaluate the effectiveness of road network classification.
The evaluation is carried out through the comparison among the metrics of a baseline model with only the traditional morphological indices and an augmented model augmented with probabilistic indices. 
More nuanced relationships between the different aspects of vitality and urban morphology are also studied.

\subsubsection{Study areas}

Considering the diversification of locations and similarity of development level, nine cities spanning multiple continents are chosen as the study areas, including Singapore, Shanghai, Beijing, New York, Chicago, Seattle, London, Paris, and Berlin.
For the convenience of data collection and analysis, each city is latticed into approx.\ 1km $\times$ 1km grids.
The grids are in accordance with the 1km resolution grids from WorldPop \citep{worldpop,Tatem.2017,Lloyd:2017et,Lloyd:2019gc}.
These grids contain their estimated population inside, which is one of our indicators of urban vitality. 
Other indicators of both urban vitality and morphology are processed to match the same grid system and level.

\subsection{Measurement of Urban morphology}
\subsubsection{Colored road hierarchy diagram}
\label{Sec.3.2.1}

A part of our method relies on OSMnx, a Python module developed by \citet{BOEING2017126}, which allows querying OpenStreetMap data in a programming environment. 
It enables a straightforward generation of monochrome road network diagrams.
This study takes a step further.
While monochrome road network diagrams represent road hierarchy through visual thickness, we propose an improved version that represents road hierarchy with both thickness and color.
We name such sort of diagram \textit{Colored Road Hierarchy Diagram (CRHD)}.
We regroup the original road classes from OpenStreetMap into condensed five: motorway, primary roads, secondary roads, tertiary roads and minor roads. 
Each class is specified a color and a thickness.
A CRHD is generated through querying the geospatial data of the road network in a given area through OSMnx and rendering the roads based on our regroup and appearance rules.
Thanks to the consistent standard of OpenStreetMap data around the world, our generation method of CRHD is universal and global.
A Python script is developed as part of this study to carry out this generation process, which we open-source.

The colors of CRHDs aid the visual representation of road hierarchy and provides an additional dimension of information to deep learning algorithms. Another merit of CRHD is the convenience of conducting color thresholding to tone down a certain grade of roads.
For instance, minor roads which are in the lightest color can be erased by truncating the pixels with high RGB values. 

\begin{figure*}[tbp]
\centering 
\subfigure[Radial (Paris).]{
\label{Fig.2.sub.1}
\includegraphics[width=0.235\textwidth]{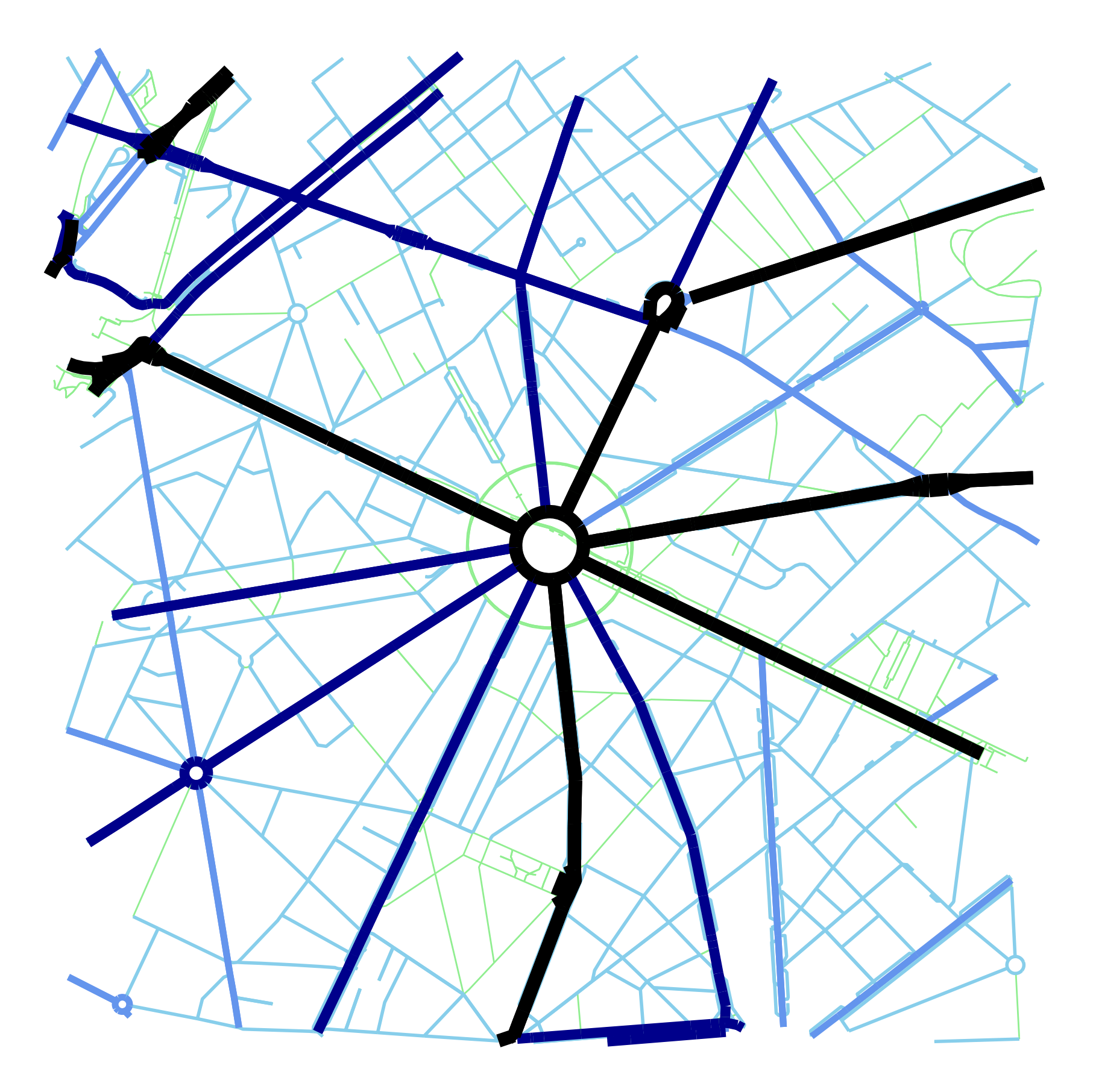}}
\subfigure[Organic (Stuttgart).]{
\label{Fig.2.sub.2}
\includegraphics[width=0.235\textwidth]{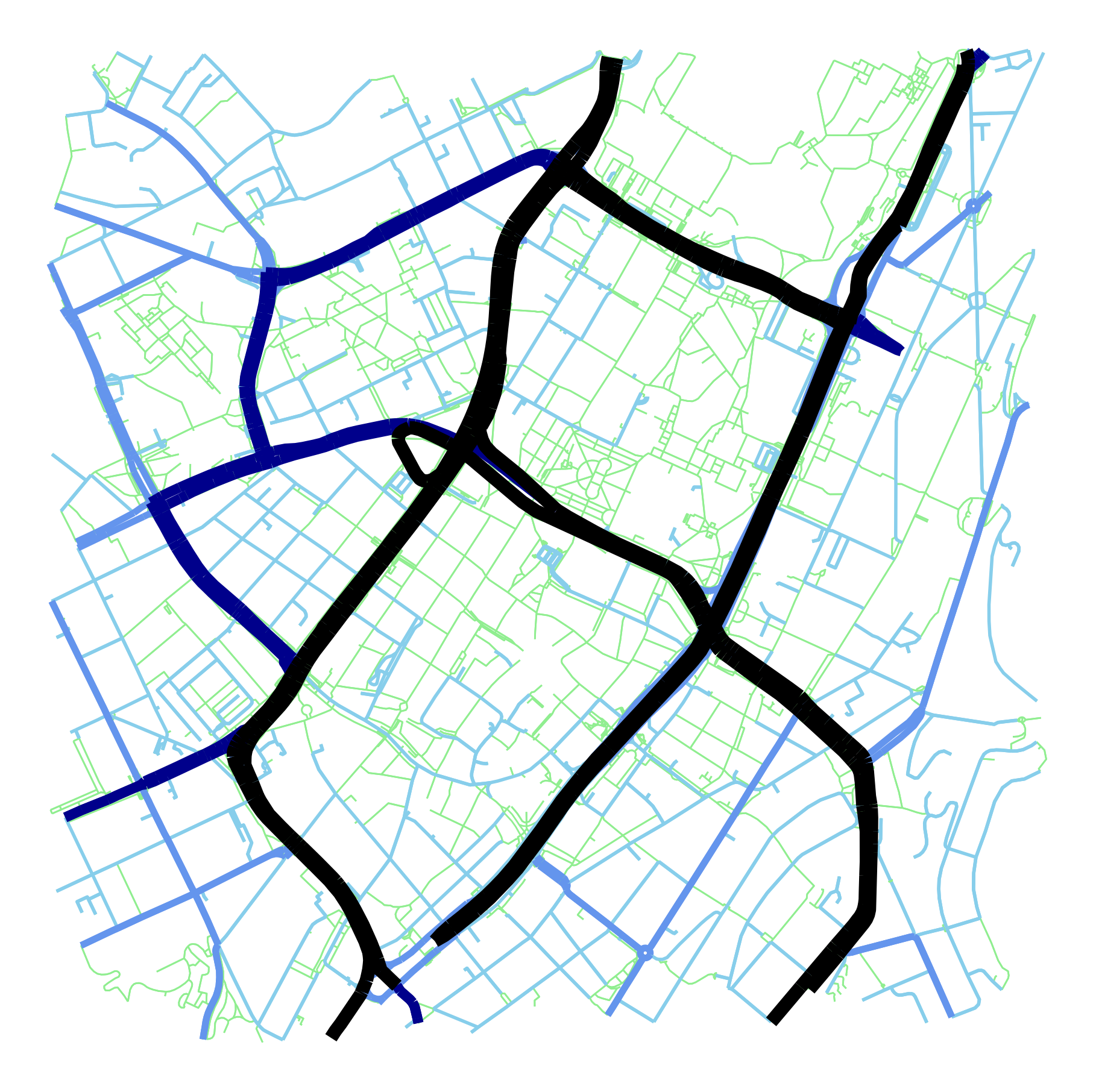}}
\subfigure[Gridiron (Chicago).]{
\label{Fig.2.sub.3}
\includegraphics[width=0.235\textwidth]{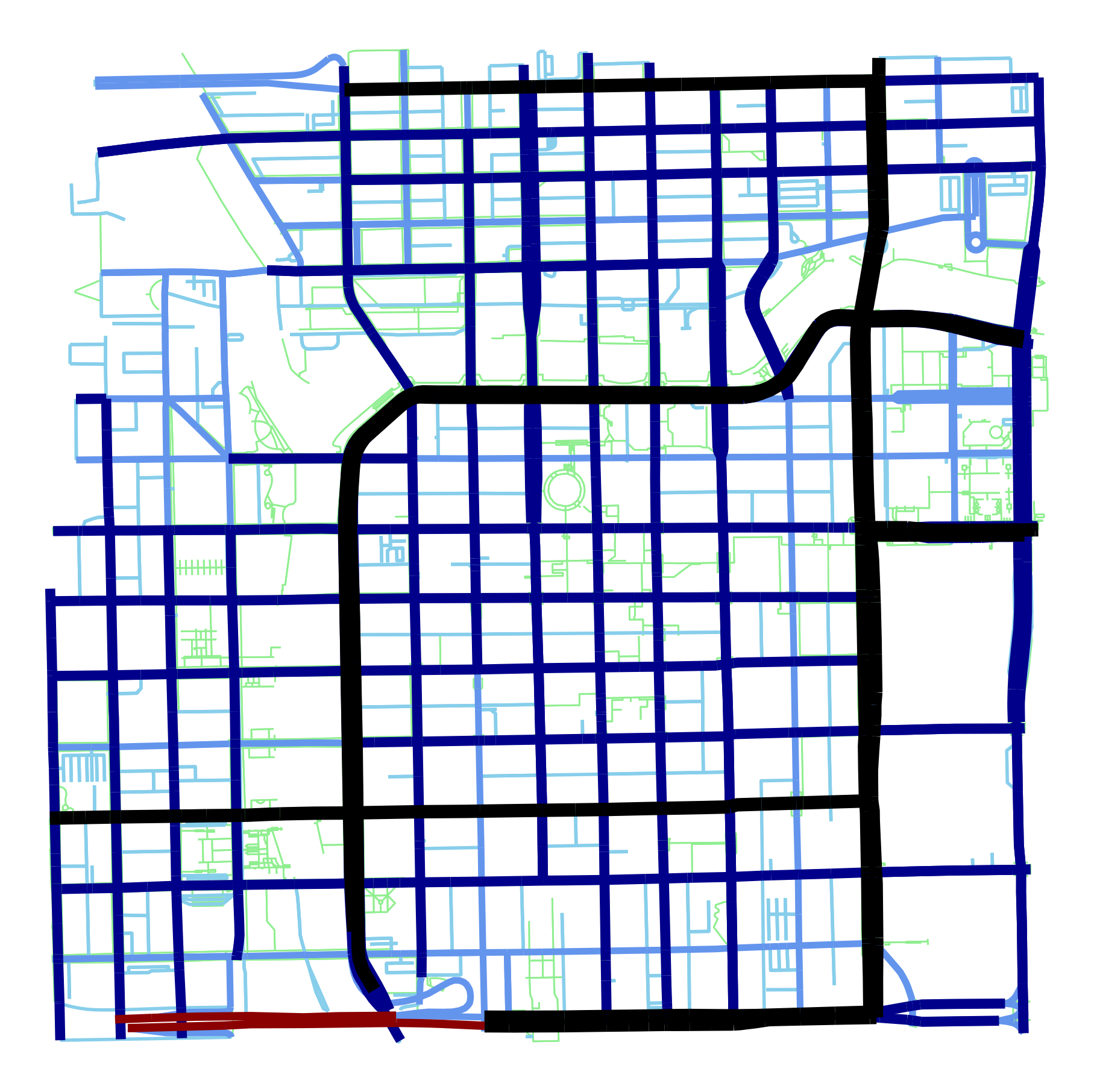}}
\subfigure[No pattern (London).]{
\label{Fig.2.sub.4}
\includegraphics[width=0.235\textwidth]{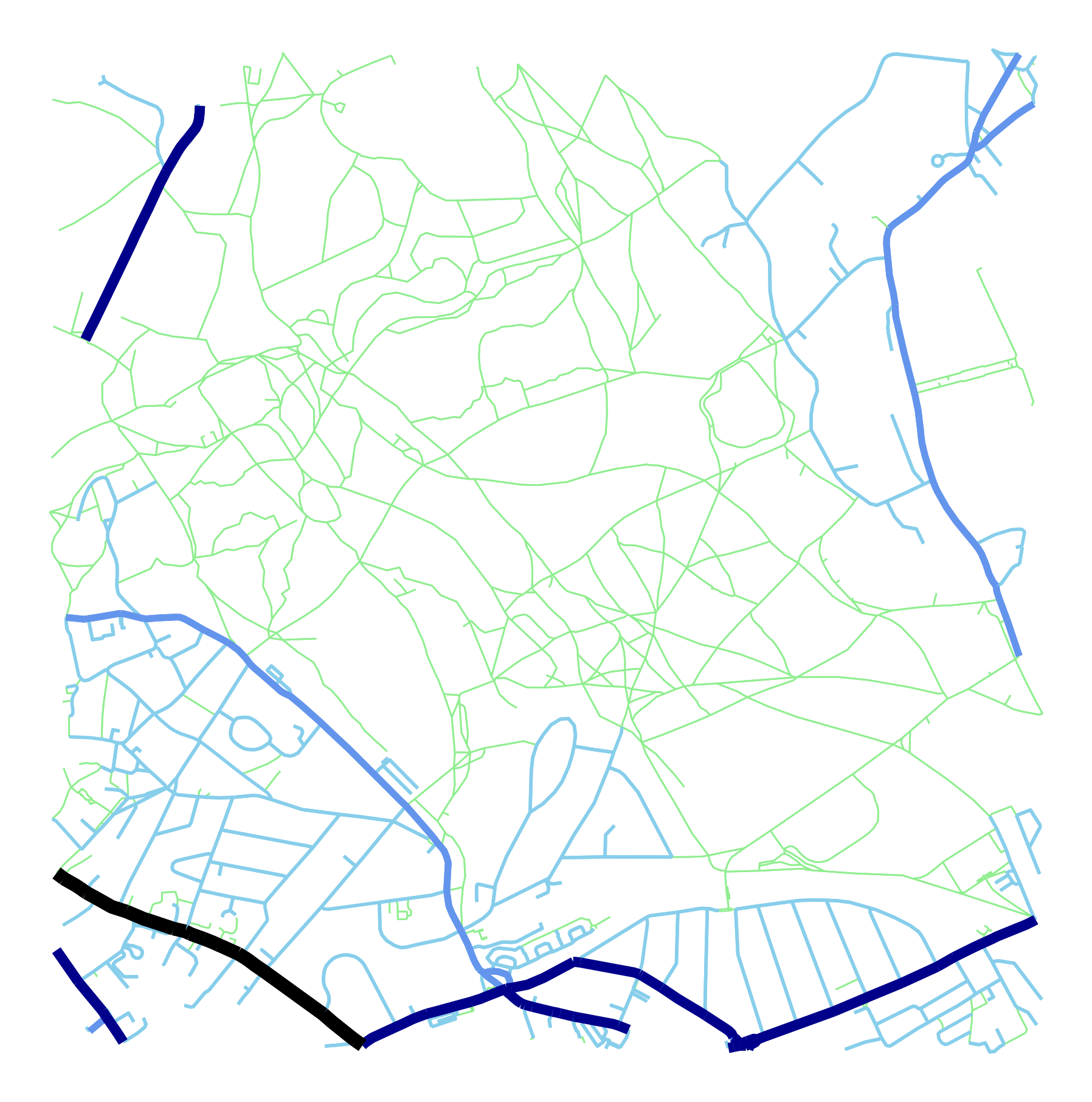}}
\caption{Quintessential CRHDs (2km $\times$ 2km) for different road network categories, as generated with our tool. The underlying data is obtained from OpenStreetMap.}
\label{Fig.2.main}
\end{figure*}

Previous studies have proposed multiple divisions of road network categories \citep{Danielle2002urban, plater2003lexicon, marshall2005urban, chan2011urban, han2020classification}.
Because our study covers cities around the world, universal suitability of the taxonomy is important.
Hence, we value the commonalities of the taxonomies. Radial, organic, and gridiron are found to be the three types that most frequently occur. Additionally, we believe that these categories are easily discernible which is beneficial to alleviate errors in manual labelling. We name these three categories ``patterned road network categories'' in comparison to another category we add --- ``no pattern''. The intention of ``no pattern'' is to avoid misclassifying instances that do not belong to none of the three patterned categories. 
As a result, the study employs four target categories.
The representative CRHDs for radial, organic, gridiron and no pattern are illustrated in Figure~\ref{Fig.2.main}. 

\subsubsection{ResNet-34}
\label{Sec.3.2.2}
ResNets, the abbreviation of Residual Neural Network, was first introduced by \citet{he2015deep}. Compared with common convolutional neural networks, ResNets reformulate the layers as learning the residual functions referring to the layer inputs. 
Compared to earlier architectures such as VGGs, ResNets offer better performance per parameter and faster inference speed \citep{DBLP}. ResNets are constructed by a series of building blocks with ``shortcuts''. In this paper, we use an improved scheme of building blocks \citep{he2016identity}. After some preliminary experiments on model selection, this study opts for ResNet-34.

The task of the model is to classify the CRHDs into the given categories. Further, the model is supposed to perceive the characteristics from a human's view and convert this information from an image to a feature vector. 
The paper employed the four probabilities exported from the softmax layer as the embedded feature vector. The numbers in the vector represent the probabilities of this CRHD belonging to each category.

\subsubsection{Road network classification}

\begin{figure*}[tbp]
\centering
\includegraphics[width=\textwidth]{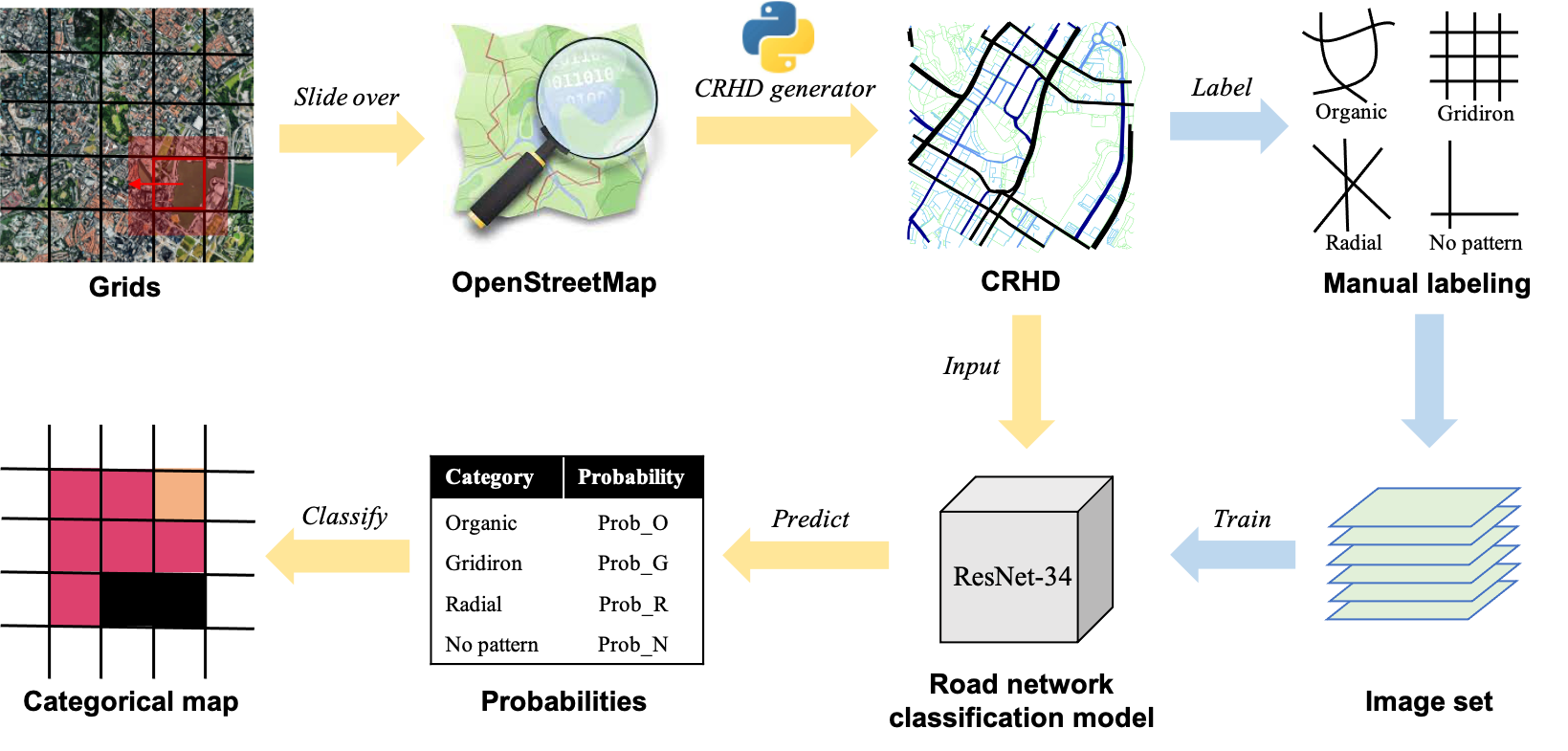}
\caption{Process of road network classification.}
\label{Fig.1.5}
\end{figure*}

Figure \ref{Fig.1.5} outlines the process of road network classification. It is implemented in two steps. The blue arrows indicate the first section, which is the training of the road network classification model. The training set include samples from our aforementioned study area (20\%) and also the other cities around the world (80\%), which are 108 cities in Asia, Europe, North America, and Africa in our case. Considering our limited capability of manual labelling, the percentage is a trade-off between being more precise in given areas and being more robust to unknown areas. We believe we choose the percentage that reconciles the model's performance in our study area with its widespread application.
Training samples must possess a quintessential road network of one of the target categories. 
The samples are fed to the road network classification model for training and validating.

The yellow arrows denote the second portion of the approach --- the process of grid road network classification. Avoiding unbuilt areas, the study excludes grids with no buildings. The CRHDs used for prediction have a radius of 1km. It is worth noting that the scale of a CRHD is twice as large as a grid. This difference in scale is under the consideration of the interaction and conjunction of road networks among the adjacent grids. It is unreasonable to take out one grid and study it separately without counting the effect of the contextual road networks. Therefore, the CRHD is concentric with the grid but with a doubled extent. 
Given an input of CRHD, the classification model would return its probability of each category. Then the CRHD would be assigned to the category with the highest probability. The probabilities could be further harnessed as numerical morphological features in quantitative studies, while the categories could be used for categorical analysis and the generation of the categorical maps for the cities.

\subsubsection{Morphological Indices}

Urban morphology usually comprises three major components: street, building, and block, which we follow in our study. We adopt the morphological indices reported correlative to urban vitality in previous studies \citep{ye2018block,yang2021elaborating,wu2019influence,bocher2018geoprocessing,li2020urban}, including road density, intersection density, building density, average building footprint area, block density, average block area, and land use mixture.
In addition to OpenStreetMap, in some cases we use official open datasets from the municipal governments.
After acquiring the shapefiles of street, building, and land use for each city, our Morphoindex generator is applied to realise the aforementioned morphological indices automatically for each grid in the city. Other than these traditional indices, the road network classification model provides four probabilistic features as well. They are added into the feature space and regarded as supplementation or augmentation of human sense, and they may also be seen as the level of intensity of a particular form. The effectiveness of the augmentation is evaluated in the following sections. The final list of morphological indices is shown in Table \ref{Tab.0} with a detailed description of each index. It is worthwhile to point out that the land use mixture in a grid is gauged by Shannon entropy \citep{shannon1948mathematical}, which describes the level of uncertainty and mixture. It could be calculated through the following equation:

\begin{equation}
LUM = -\sum\limits_{i=1}^np_i\log p_i\label{eq:2}
\end{equation}
where $n$ denotes the total amount of land use category in the grid, and $p_i$ is the proportion of the area of $i$th category.

\begin{table*}
\centering
\small
\caption{Morphological indices incorporated in our study.}
\label{Tab.0}
\renewcommand\arraystretch{1.5}
\begin{tabularx}{\textwidth}{llXX}
\toprule
\textbf{Component}&\textbf{ID}&\textbf{Name}&\textbf{Description}\\
\midrule
Street&Prob\_R&Probability of being radial&Road network's probability of being radial\\
&Prob\_O&Probability of being organic&Road network's probability of being organic\\
&Prob\_G&Probability of being gridiron&Road network's probability of being gridiron\\
&Prob\_N&Probability of being no pattern&Road network's probability of being no pattern\\
&RD&Road density&The total length of roads in a grid divided by the area of the grid\\
&InD&Intersection density&The total number of intersections in a grid divided by the area of the grid\\
\midrule
Building&BuD&Building density&The total area of building footprints in a grid divided by the area of the grid\\
&ABFA&Average building footprint area&The average area of building footprints in a grid\\
\midrule
Block&BlD&Block density&The total area of blocks in a grid divided by the area of the grid\\
&ABA&Average block area&Average area of blocks in a grid\\
&LUM&Land use mixture&Shannon entropy of land use in the grid\\
\bottomrule
\end{tabularx}
\end{table*}

\subsection{Measurement of urban vitality}

\begin{table*}
\centering
\small
\caption{Urban vitality indicators and data sources.}
\label{Tab.1}
\renewcommand\arraystretch{1.5}
\begin{tabularx}{\textwidth}{lXXX}
\toprule
 Aspect&Indicator&Calculation&Data source\\
\midrule 
Built environment vibrancy&POI density&Average kernel density of POIs&OpenStreetMap\\
Human activity density&Tweet density&Count of tweets&\href{https://developer.twitter.com/en}{Twitter Developer API}\\
Nighttime vibrancy& NTL brightness&Monthly average NTL brightness&\href{https://ngdc.noaa.gov/eog/viirs/}{VIIRS DNB}\\
Population density&Population size&Estimated population&\href{https://https://www.worldpop.org/}{WorldPop}\\
Tourism vibrancy&Airbnb listing density&Average kernel density of Airbnb listing&\href{http://insideairbnb.com/about.html}{Inside Airbnb}\\
\bottomrule
\end{tabularx}
\end{table*}

This study inherits the frequently-used aspects of urban vitality from existing studies, including built environment vibrancy, human activity density, and nighttime light brightness. Two extra aspects of urban vitality, population density and tourism vibrancy are supplemented. Table~\ref{Tab.1} shows the inventory of our urban vitality aspects, indicators and data sources.
Provided that each of those vitality indicators reflects only one characteristic of urban vitality (human --- tweet density, population; environment --- POI density, NTL, and Airbnb density), we define an indicator called ``vitality score'' that evaluates the level of interactions (among humans and between human and environment) by integrating our five vitality aspects. Vitality score ranges from 0 to 100. It is calculated by three steps: (1) standardize the vitality indicators, (2) sum them up, and (3) rescale.

\subsection{LightGBM}
LightGBM \citep{ke2017lightgbm} is a prevailing machine learning method in the industry, thanks to its fast speed, high accuracy, and ability to filter important features.
It is evolved from gradient boosting decision tree (GBDT).
Given a training dataset $\{(x_1,y_1),(x_2,y_2), \cdots,(x_n,y_n)\}$, for each epoch, GBDT aims to find the best approximation $\hat{f}$ to minimize the expectation of the loss function $L(y,f(x))$ (Eq. \ref{eq:3}). 

\begin{equation}
\hat{f}=\mathop{\arg\min}\limits_{f}{E_{x,y}[L(y,f(x))]} \label{eq:3}
\end{equation}

Based on the idea of GBDT, LightGBM utilizes two new techniques, Exclusive Feature Bundling (EFB) and Gradient-based One-Side Sampling (GOSS), to accelerate the training process and improve degree of accuracy.


In this study, we harnessed the Python package, lightgbm, to obtain an easy-to-use application of the LightGBM model.
As a tree-based model, LightGBM is capable of ranking feature importance. The principle behind it roots from the process of the generating decision trees. The importance of a feature could be evaluated through its total number of times being selected for treenode split.

\section{Results}
\label{Sec.4}
\subsection{Road network classification}
\label{Sec.4.1}
The ResNet-34 was trained based on the labelled image set with 521 images, including 136 gridiron samples, 163 organic samples, 82 images radial samples, and 140 no pattern samples. The image set was randomly split into a training set with 432 images, a validation set with 49 images and a testing set with 40 images (10 images for each category). We truncated the minor roads (service roads, residential roads, etc.) from CRHDs to eliminate noise. This method was proven to be useful in our trials. After parameter tuning, the learning rate, batch size, and number of epochs was set to 0.0005, 2, and 30 respectively. The overall accuracy achieved in the testing set was 0.875. In contrast, when using monochrome diagrams, the accuracy dropped to 0.75, affirming the value of colored diagrams. 

\begin{figure*}[tbp]
\centering 
\subfigure[Confusion matrix]{
\label{Fig.5.sub.1}
\includegraphics[width=0.49\textwidth]{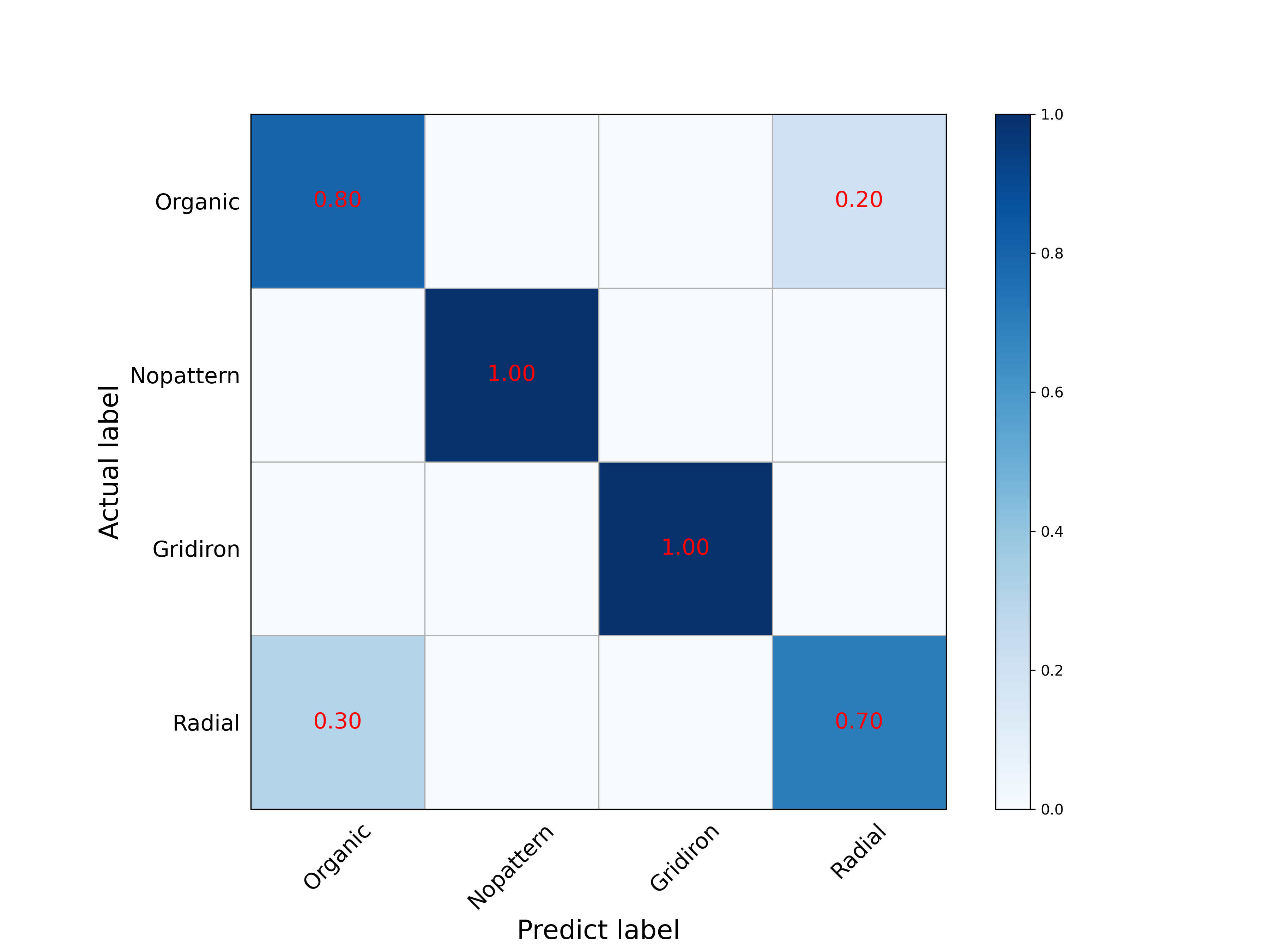}}
\subfigure[ROC curves]{
\label{Fig.5.sub.2}
\includegraphics[width=0.48\textwidth]{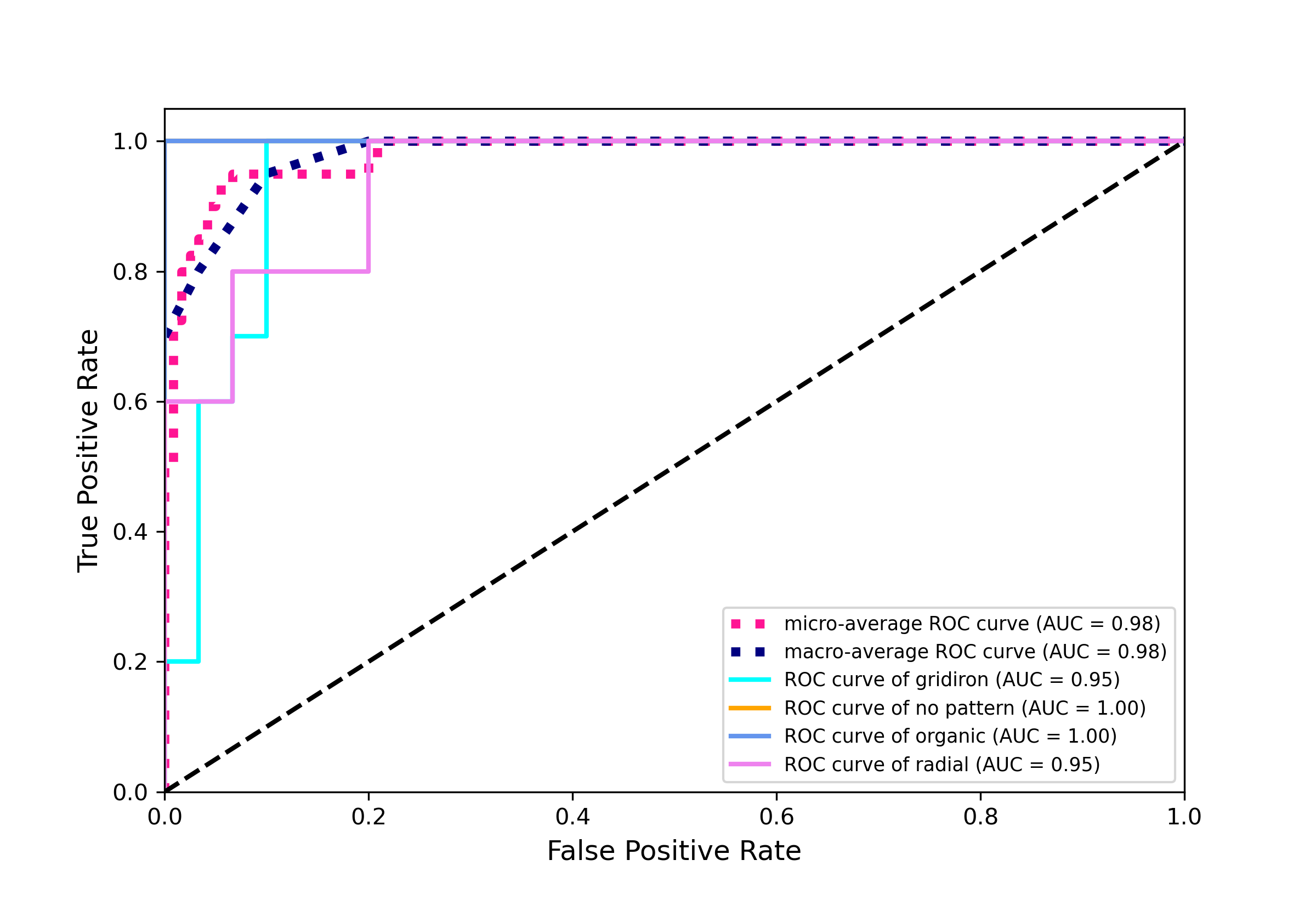}}
\caption{Metrics for the road network classification model.}
\label{Fig.5.main}
\end{figure*}

Figure \ref{Fig.5.main} provides more detailed metrics such as the confusion matrix and the ROC curves of the classification model. The diagonal of the confusion matrix (Figure \ref{Fig.5.sub.1}) denotes the accuracy of classification for each category. The model performed the best in recognizing no pattern road networks and gridiron road networks with a hundred percent accuracy. The accuracy of classifying organic and radial road networks are relatively lower which are 0.8 and 0.7, respectively. There are possible explanations for this outcome. First, the classification of road networks relies heavily on the skeleton of major roads as they are easier for the model to interpret. It is more advantageous for those categories that are consistent in characteristics of major roads within the group such as gridiron and no pattern. The major roads of the gridirons are in most cases consistent in being straight and orthogonal, and no pattern road networks have rarely major roads. In comparison, organic road networks have the winding and intertwined major roads which create large blocks with irregular shapes, making it difficult to cover all kinds of samples. The characteristics of major roads of radial network are evident but very sensitive to the position of the radial center in the image. The pattern of radial is distinct only if the radial center locates near the center of the image. However, the grids from WorldPop usually cannot capture the radial pattern properly, causing a decline in the accuracy. Second, there is an unbalance for the sample size of the four categories in the image set. Radial images were fewer than the counterparts due to fewer typical cases. This disparity also contributes to the final difference in accuracy. Although discrepancy of accuracy among categories was found in the confusion matrix, the AUC values did not differ a lot and were all above 0.9 (Figure \ref{Fig.5.sub.2}), indicating acceptable classification errors for all the categories. Only the ROC curve of radial was slightly under the average ROC curve. However, the model has been satisfactory enough.

\begin{figure*}[htbp]
\centering 

\makebox[\linewidth][c]{%
\subfigure[Seattle]{
\label{Fig.6.sub.1}
\includegraphics[width=0.44\textwidth]{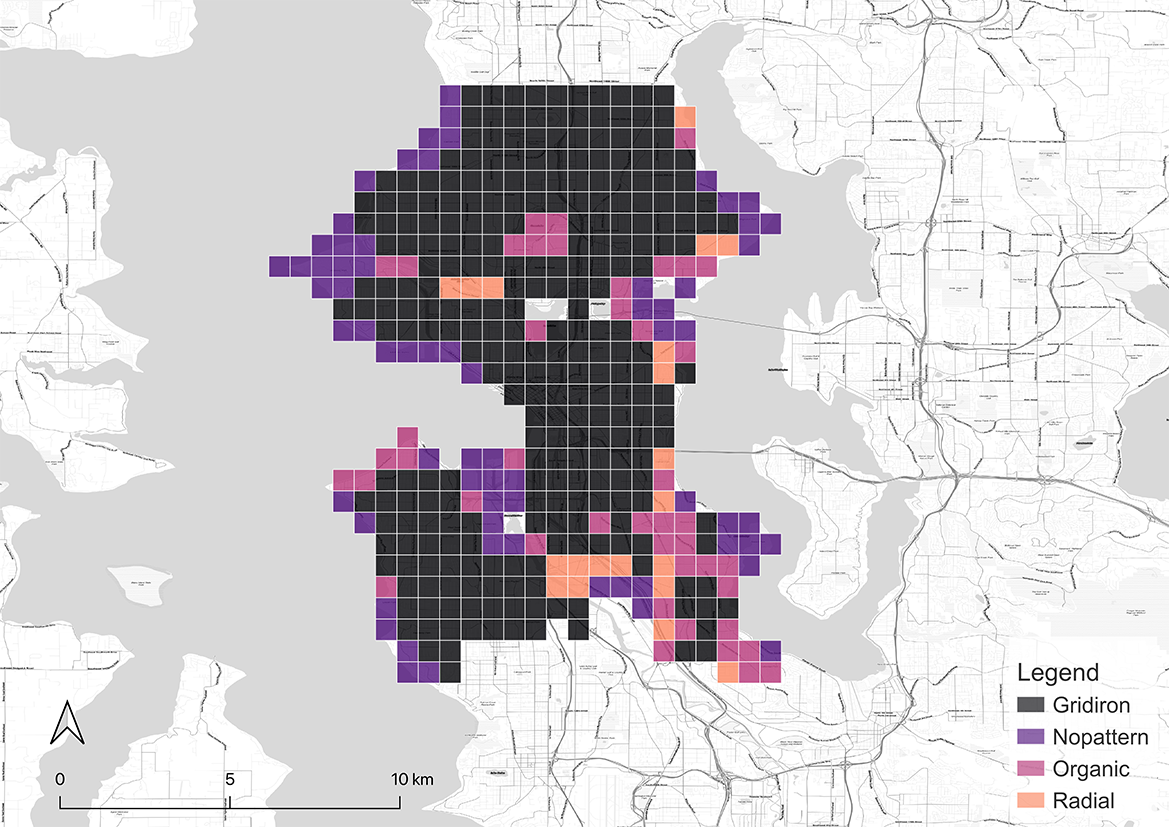}}
\subfigure[Chicago]{
\label{Fig.6.sub.2}
\includegraphics[width=0.44\textwidth]{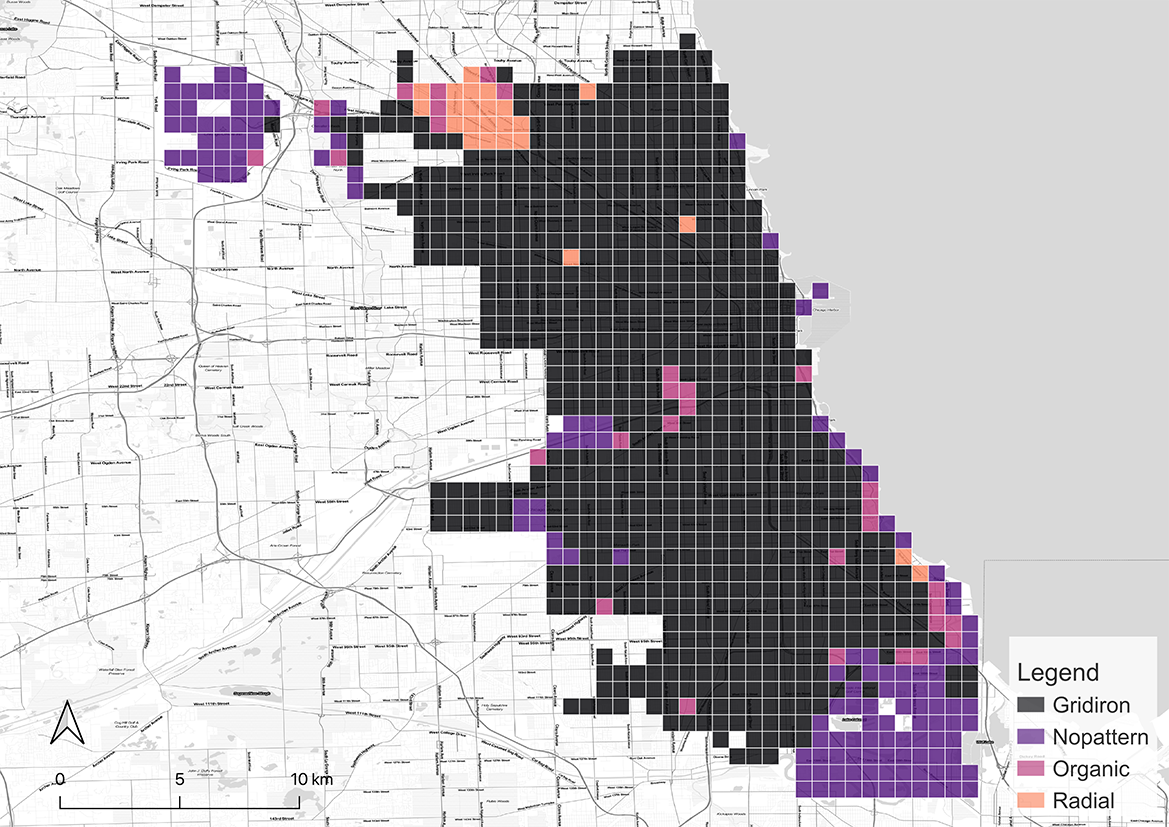}}
\subfigure[London]{
\label{Fig.6.sub.3}
\includegraphics[width=0.44\textwidth]{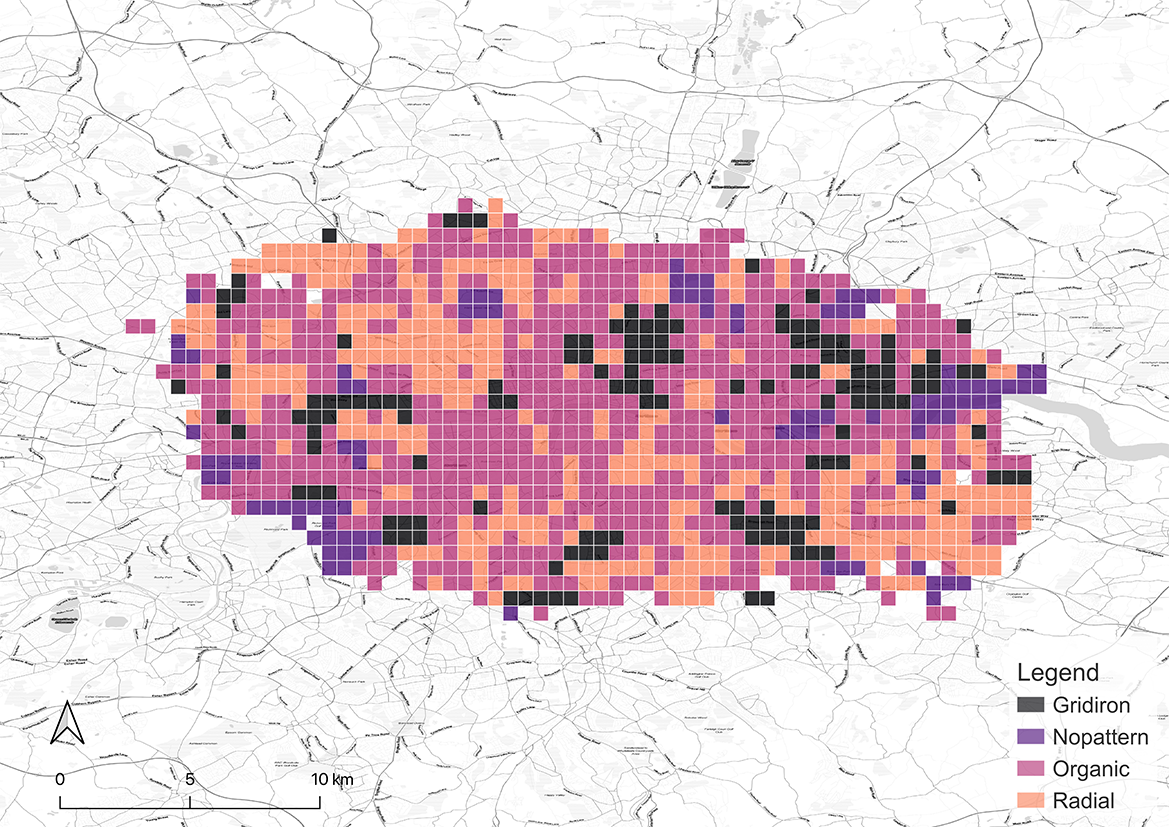}}
}
\makebox[\linewidth][c]{%
\subfigure[Paris]{
\label{Fig.6.sub.4}
\includegraphics[width=0.44\textwidth]{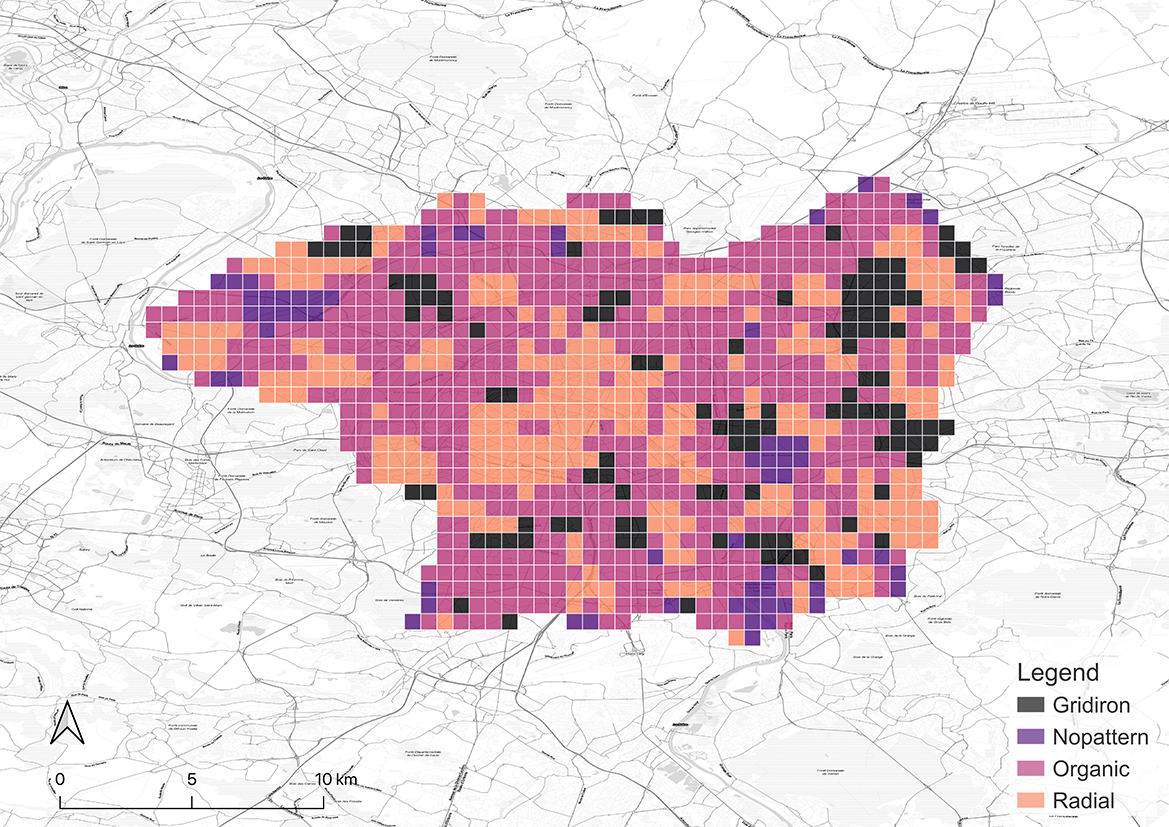}}
\subfigure[Berlin]{
\label{Fig.6.sub.5}
\includegraphics[width=0.44\textwidth]{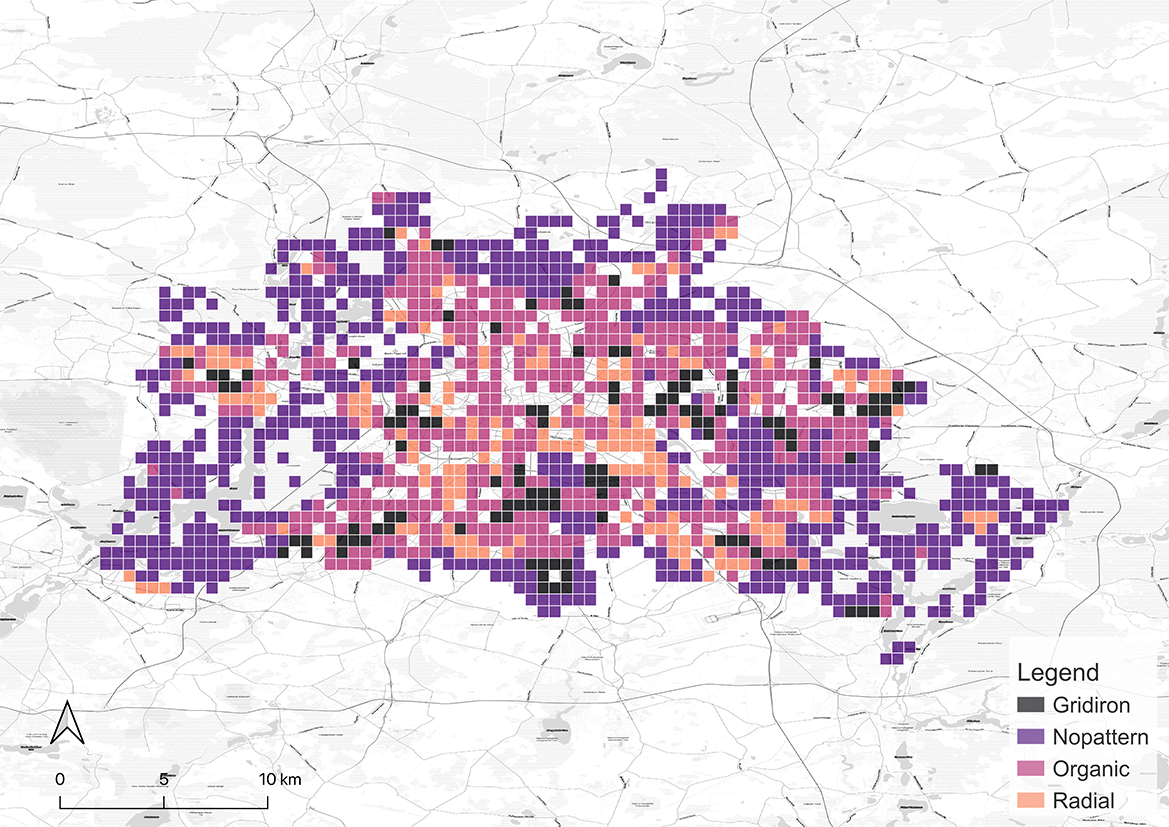}}
\subfigure[Singapore]{
\label{Fig.6.sub.6}
\includegraphics[width=0.44\textwidth]{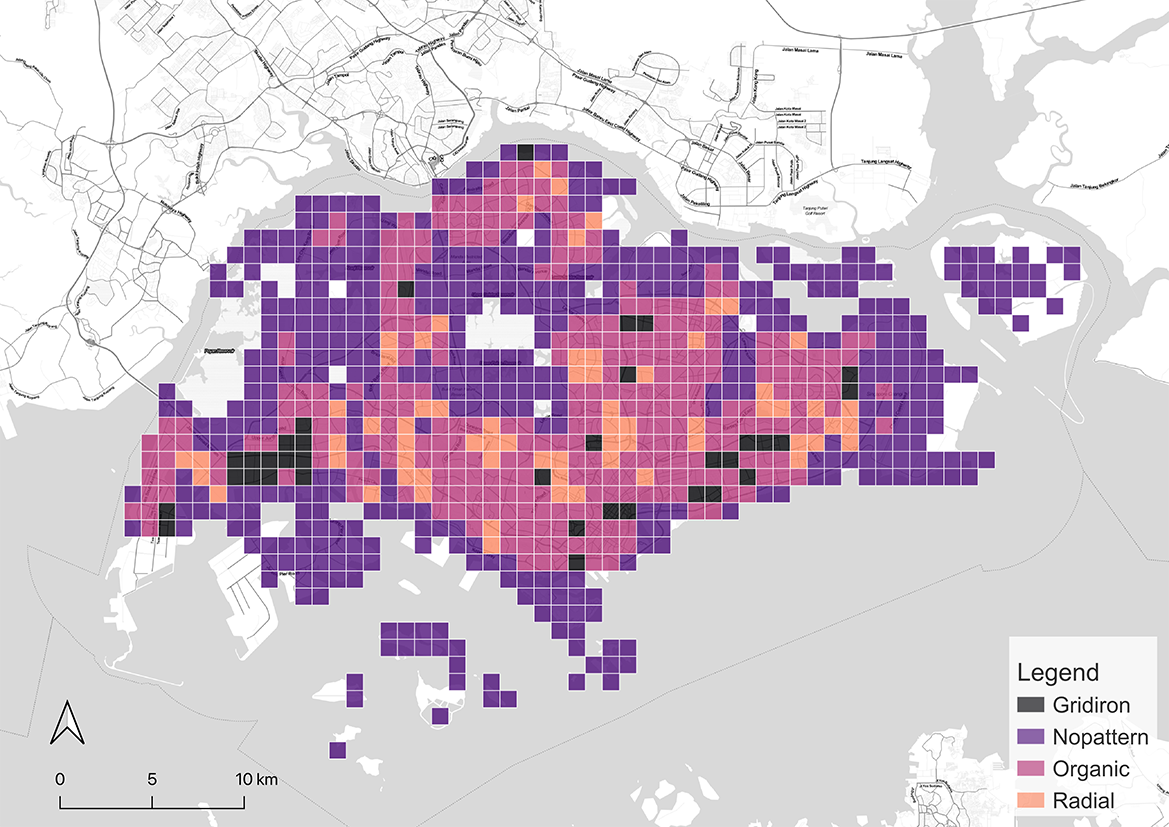}}
}
\makebox[\linewidth][c]{%
\subfigure[New York]{
\label{Fig.6.sub.7}
\includegraphics[width=0.44\textwidth]{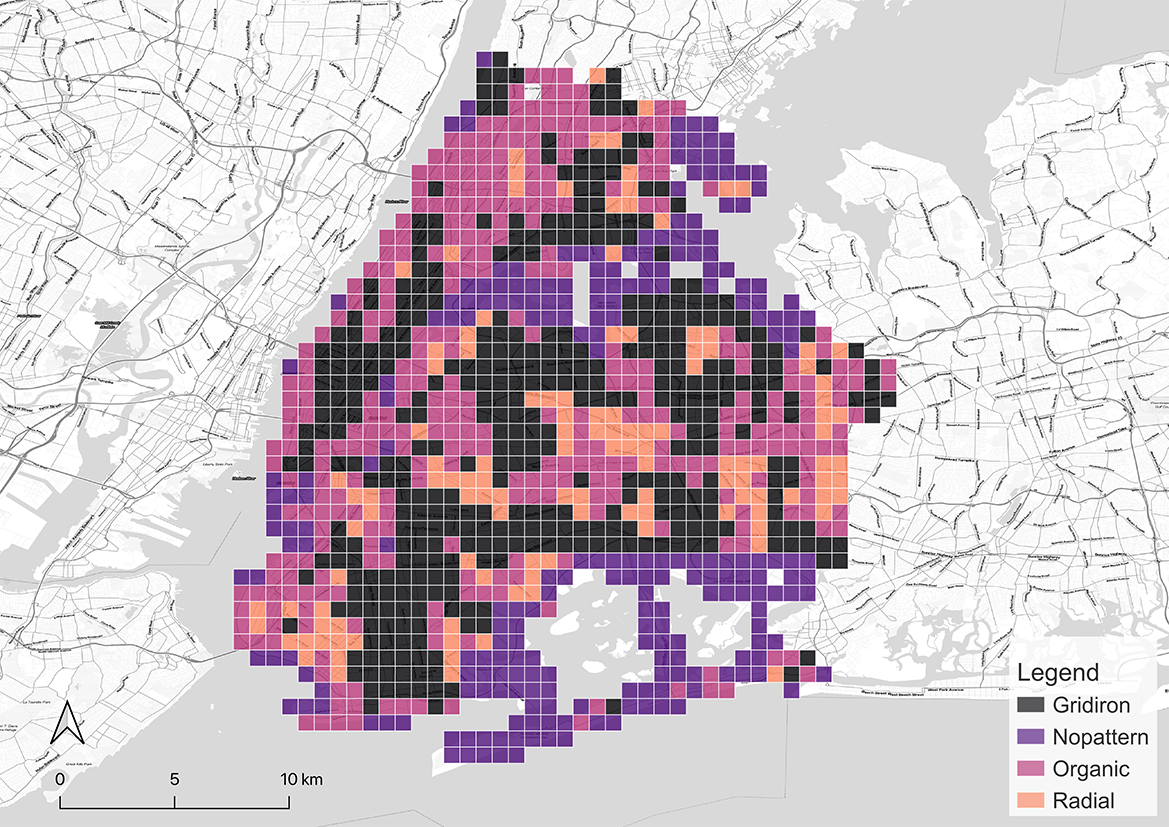}}
\subfigure[Beijing]{
\label{Fig.6.sub.8}
\includegraphics[width=0.44\textwidth]{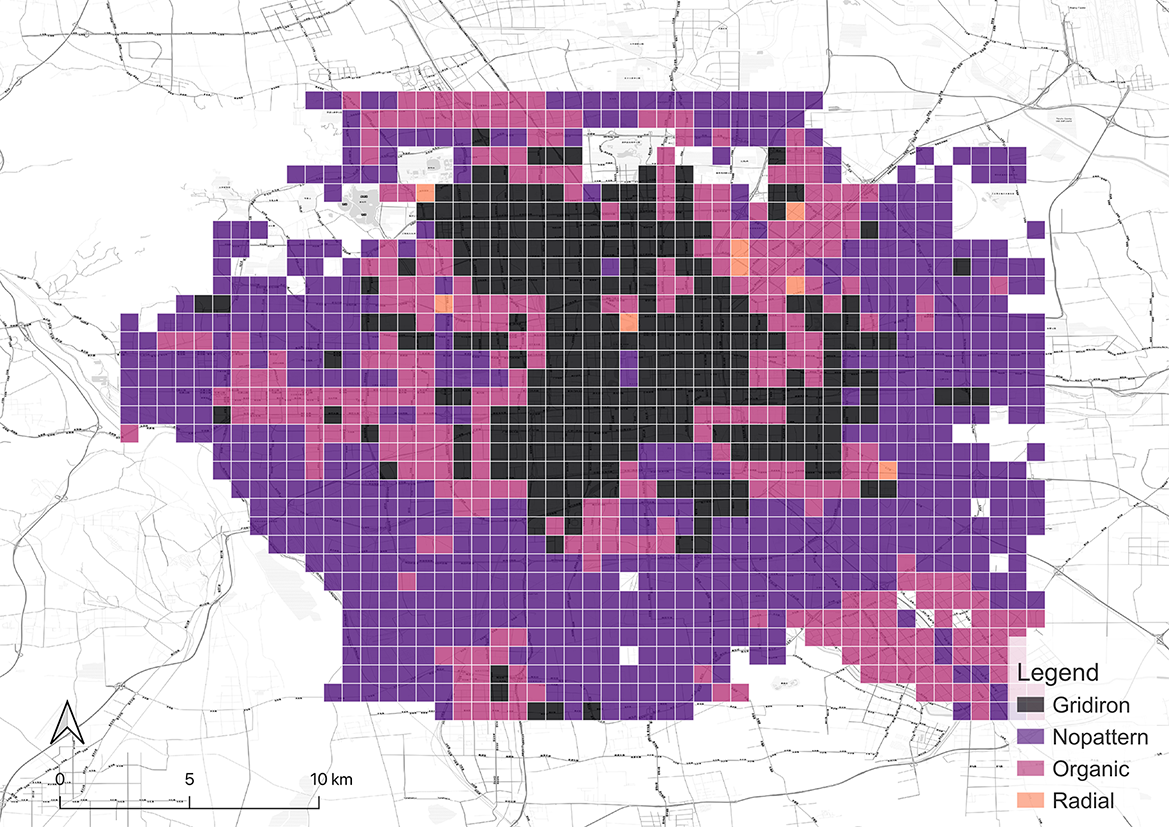}}
\subfigure[Shanghai]{
\label{Fig.6.sub.9}
\includegraphics[width=0.44\textwidth]{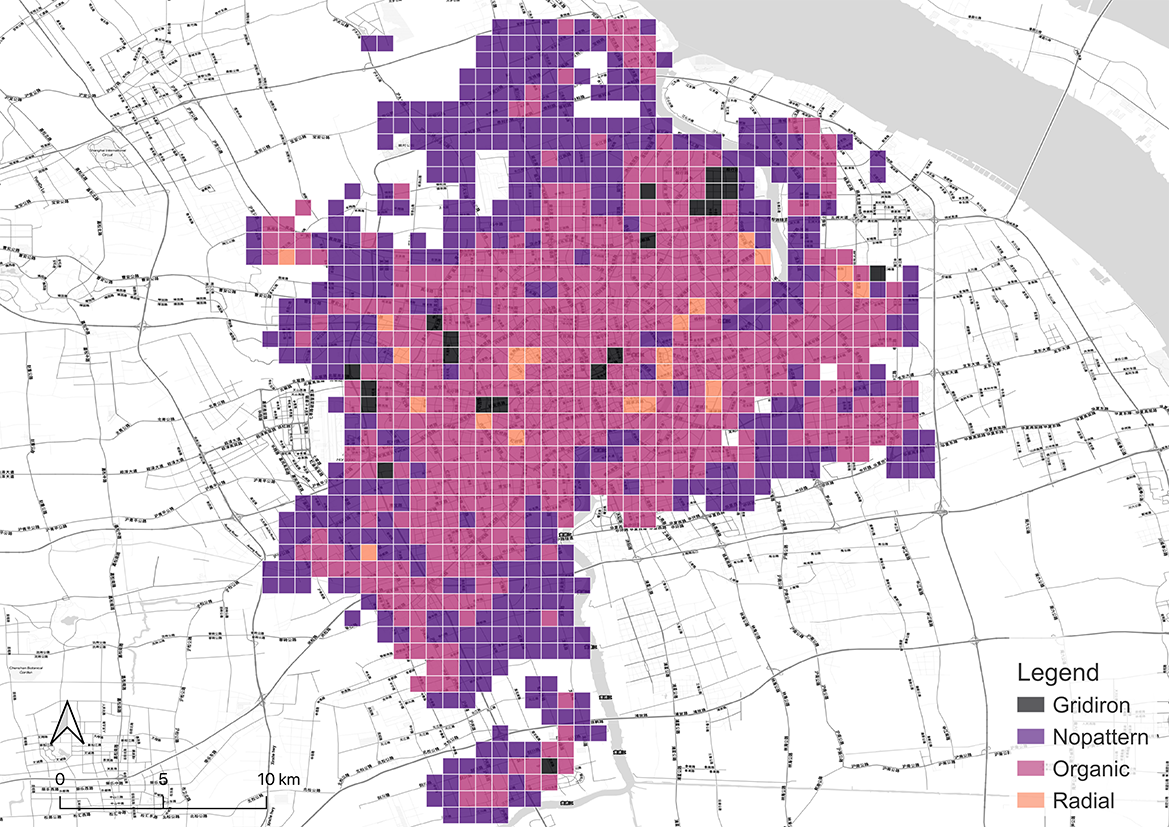}}
}
\caption{Categorical maps of the cities derived from our classification method.}
\label{Fig.6.main}
\end{figure*}

After the model is all set, it is applied to classify the grids in all the cities in our focus.
The results of the classification are shown in Figure \ref{Fig.6.main}.
They are affirmed by well-known facts, e.g.\ gridiron road network is a distinct common ground for North American cities \citep{southworth1995street, southworth2013streets}. Thus, locations in the US we selected, such as New York City, Seattle, and Chicago have most of the grids identified as gridiron. The last two are especially dominated by the gridiron pattern. New York City has obvious gridiron road networks in the downtown areas such as Manhattan, Queens, and Brooklyn while the outskirts are organic instead. Beijing fits a similar pattern. In contrast, a great proportion of the downtown area in Paris is classified as radial in accordance with the famous Parisian radial road structure, represented by the road network radiating from the Triumphal Arch. London also has a considerable number of radial grids, but they mainly distribute in the residential area outside the downtown core which is different from the setting in Paris. It is worth noting that it is the organic pattern that takes the largest portion in London and even in Paris. Another European city, Berlin, exhibits the same structure. Akin to the preference on gridiron in North America, a commonality of pattern in Europe seems to exist as well. Finally, the remaining two Asian cities, Singapore and Shanghai, are predominantly organic-dominated.

\begin{figure}[htbp]
\centering
\includegraphics[width=0.7\textwidth]{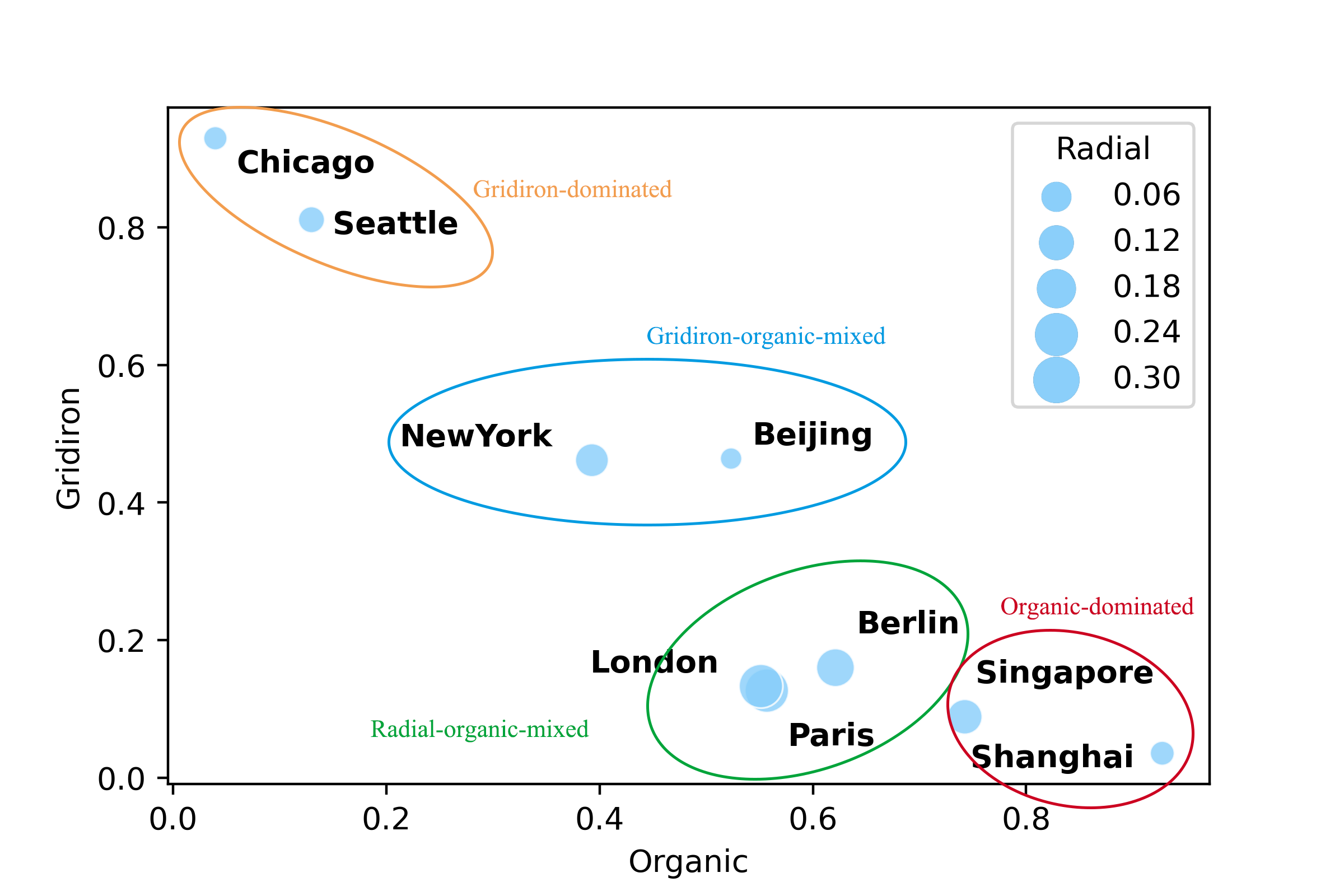}
\caption{Our classified results reveal clusters of urban form among the nine cities included in the study, where the abscissa indicates the percentage of organic grids, the ordinate indicates the percentage of gridiron instances, and the size of the point indicates the share of those that are radial.}
\label{Fig.7}
\end{figure}

Furthering the interpretation of the results, based on the commonalities of the proportion of road network categories, the nine cities are clustered into four subgroups: gridiron-dominated city (Chicago and Seattle), gridiron-organic-mixed city (New York and Beijing), radial-organic-mixed city (London, Paris, and Berlin), and organic-dominated city (Singapore and Shanghai). We can distinguish the subgroups as well in the scatter plot (Figure \ref{Fig.7}).
It should be noted that the number of no pattern grids detected in a city is highly influenced by how much outskirt region of that city is included. Considering the varying proportion of outskirts included among the cities, we exclude the grids with no pattern in this analysis to prevent such variation from disturbing the result.
This analysis may be scaled to determine automatically the predominant urban form of thousands of cities around the world.

\subsection{Effectiveness of road network classification in predicting urban vitality}
\label{sec.4.2}

In this section, we proceed to assess the effectiveness of road network classification in predicting urban vitality. 
The comparative assessment is conducted by comparing the error metrics between a baseline model trained with only traditional morphological indices and an augmented model trained with both traditional morphological indices and probabilities of road network categories. In this study, we fine-tuned every LightGBM model through a grid search technique with a 5-fold cross-validation.

\begin{table*}[htbp]
\centering
\small
\caption{R-squared of vitality indicators obtained from two standardization strategies (standardization before and after merging).}
\label{Tab.2}
\renewcommand\arraystretch{1.5}
\begin{tabular*}{\textwidth}{lrrrrr}
\toprule
  &POI density& Tweet density& NTL & Population & Airbnb Density\\
\midrule 
After merging& 0.1270& -0.0762&0.1005&-0.0789&-0.2443\\
Before merging& 0.5169& 0.0202&0.3969&0.4394&0.2854\\
\bottomrule
\end{tabular*}
\end{table*}

Before the training, standardization is necessary to handle the problem of outliers and disparity of scale among the features. Since grids from different cities need to be integrated, two standardization strategies are considered. The first strategy follows the normal sequence, which is standardizing after the merging of data, while the second one standardizes the data within each city before merging them together. LightGBMs of each vitality indicator were trained in both strategies for comparison. The results are shown in Table~\ref{Tab.2}. We witness a substantial improvement of R-squared when we standardize before merging. The R-squared of tweet density turned out to be low. Given that another study reported a moderate association between urban morphology and tweet density \citep{crooks2016user}, we attribute the contrasting result derived in our study to the underdeveloped quality of the tweet data.
We thus exclude tweet density in the calculation of the vitality score.
In another experiment, we discover higher R-squared when targets are standardized as well. In the following experiments, the study would conform to those standardization strategy.

\begin{table}[htbp]
\centering
\small
\caption{Comparison of error metrics between baseline model and augmented model, revealing the benefits of our deep learning-based approach.}
\label{Tab.3}
\renewcommand\arraystretch{1.5}
\begin{tabular*}{8cm}{lrrr}
\toprule
  &R squared& RMSE& MAE\\
\midrule 
Baseline model& 0.5957& 6.2139&4.0797\\
Augmented model& 0.6227& 6.0106&3.9611\\
\bottomrule
\end{tabular*}
\end{table}

The baseline model and the augmented model were then trained to fit the vitality score based on their feature sets. The error metrics of the two models are shown in Table~\ref{Tab.3}. In contrast to the baseline model, 
there is a comprehensive improvement on all the metrics for the augmented model. Despite not being significant, it nevertheless indicates a positive effect of the data augmentation provided by the road network classification in terms of urban vitality prediction. We adopt the augmented model in the following context.

\subsection{Differences of morphology-vitality relationships among subgroups}
\label{sec.4.3}

\begin{figure*}[htbp]
\centering
\includegraphics[width=0.95\textwidth]{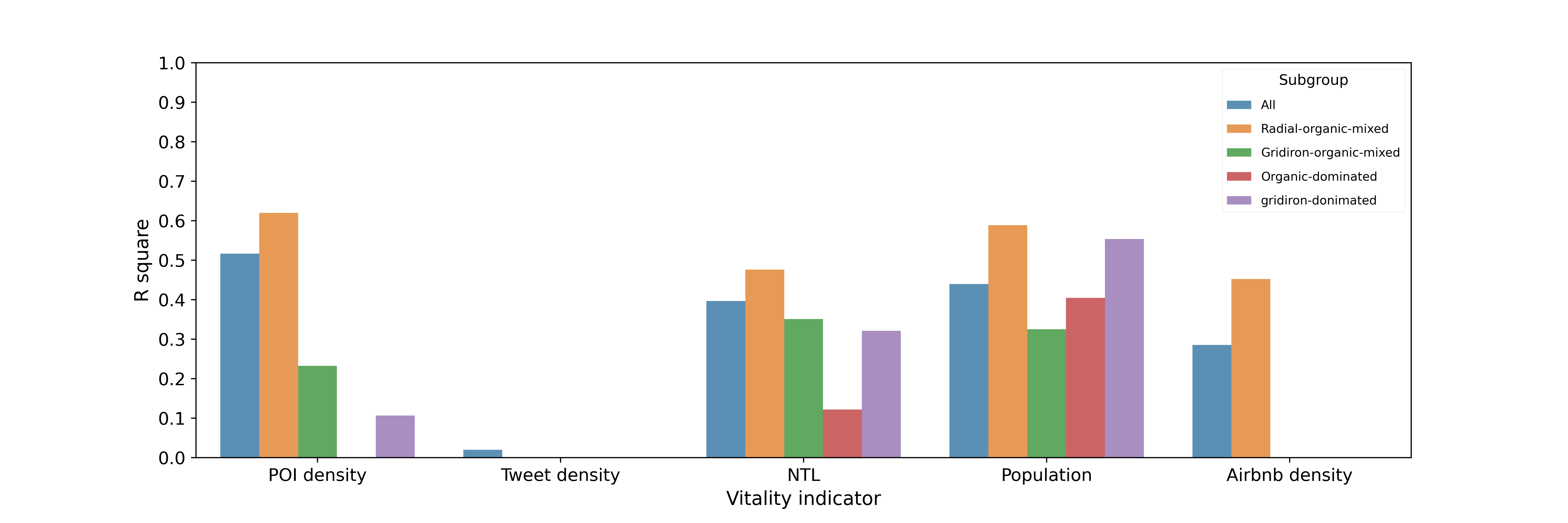}
\caption{R-squared of vitality indicators for each subgroup.}
\label{Fig.8}
\end{figure*}

This section aims to explore the possible differences of the relationship between urban morphology and vitality among the subgroups detected in Section~\ref{Sec.4.1}. A model involving all the studied cities was established in the first place to cast an overall insight into the morphology-vitality relationship. The obtained R-squared for each vitality indicator is visualized in Figure \ref{Fig.8} (the blue bars). To better explain the results, the study sets a criterion here. The relationships with the R-squared larger than 0.5, from 0.25 to 0.5, from 0 to 0.25 and equal to 0 (R-squared less than 0 would be reset to 0) would be considered as high-level, medium-level, low-level, and not relative, respectively. The results reveal that, in an overall context, urban morphology has a high-level relationship with POI density, a medium-level relationship with NTL, population and Airbnb density, but a very low-level relationship with Tweet density.

To discover the potential discrepancy between different subgroups, LightGBM models for each subgroup were trained individually in the same process as the overall model. The R-squared for each subgroup is plotted in the bar chart (Figure \ref{Fig.8}). The R-squared of tweet density for all the groups are low, which is consistent with the observation before. If we exclude this exception, we could observe the variation of the degree of morphology-vitality relationship among the subgroups. The radial-organic-mixed cities surpass the average R-squared and almost reached a high-level relationship in all aspects of vitality. 
However, gridiron-organic-mixed cities are on the contrary. They achieve lower R-squared in all the fields than the overall situation but still maintain medium-level relationships in most fields except Airbnb density. Organic-dominated cities are only detected a medium-level relationship with population with no other distinct relationship. Gridiron-dominated cities are also highly associated with population, and in the meantime have a medium-level relationship with NTL and a low-level relationship with POI density. The difference among subgroups is substantial. It unveils that cities in different subgroups (with different combination of patterned road network categories) are likely to possess different morphology-vitality relationship. 

\subsection{Relationship between urban morphology and vitality at the grid scale}
\label{sec.4.4}

\begin{figure*}[htbp]
\centering 
\subfigure[POI density.]{
\label{Fig.9.sub.1}
\includegraphics[width=0.32\textwidth]{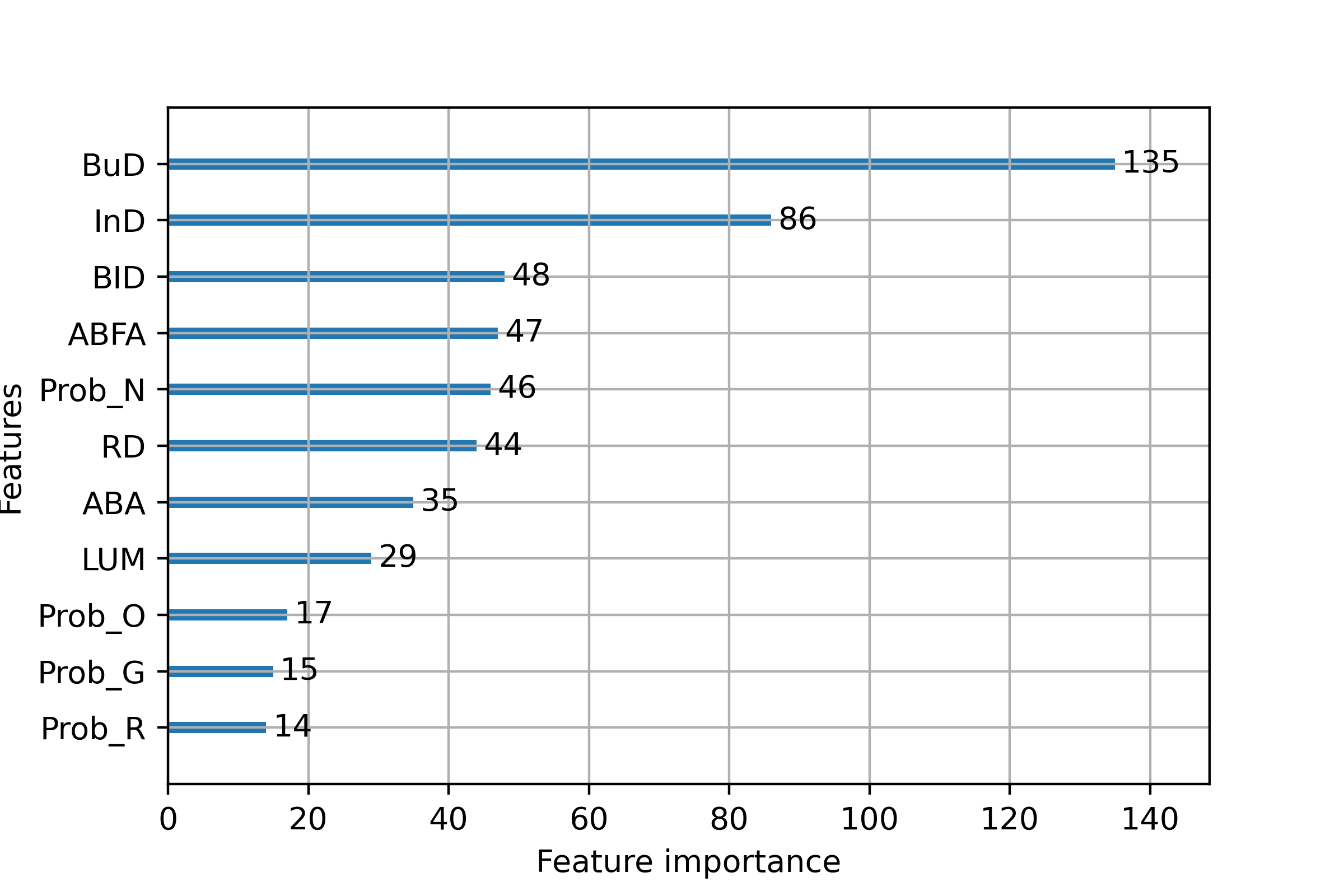}}
\subfigure[Tweet density.]{
\label{Fig.9.sub.2}
\includegraphics[width=0.32\textwidth]{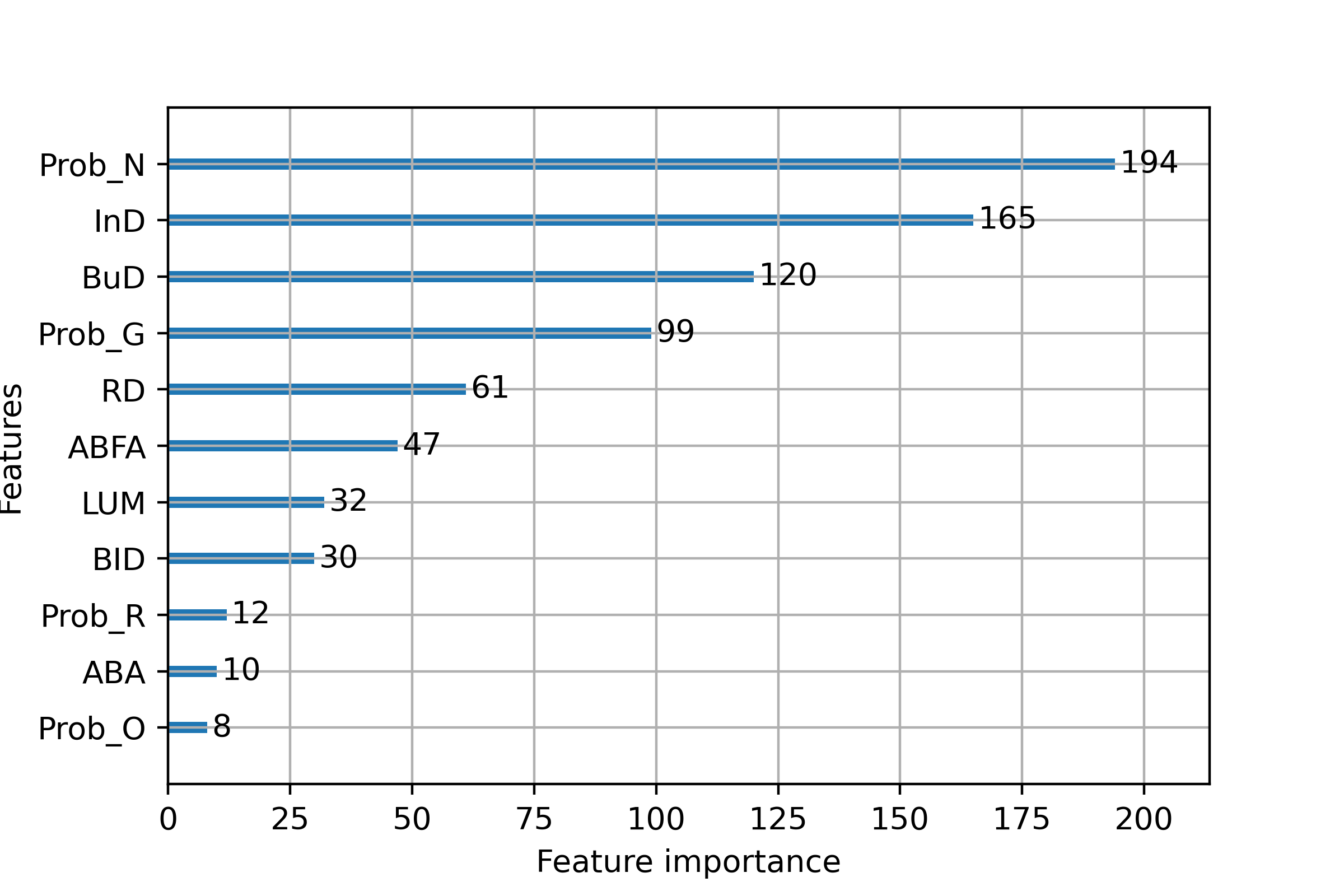}}
\subfigure[NTL.]{
\label{Fig.9.sub.3}
\includegraphics[width=0.32\textwidth]{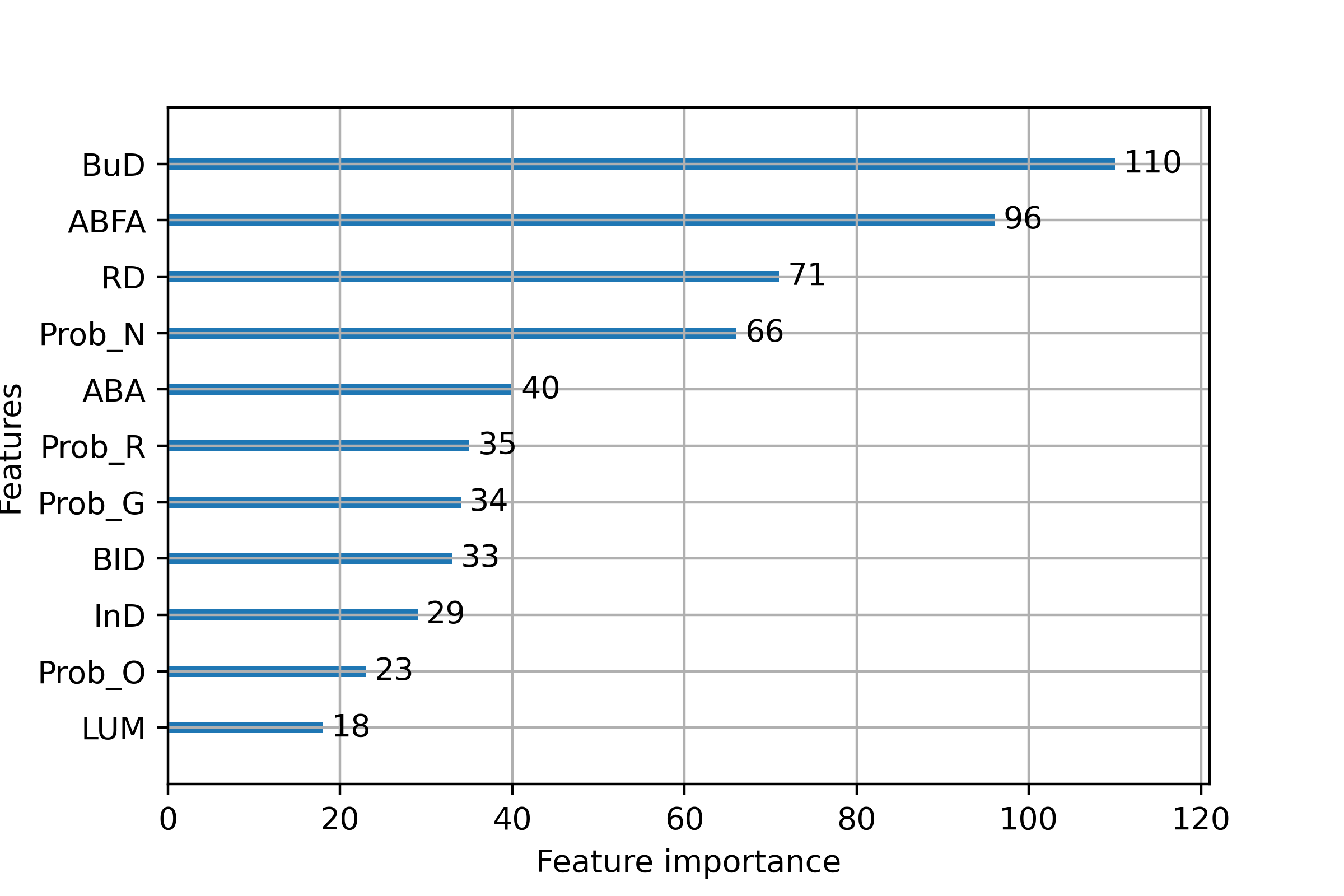}}
\subfigure[Population.]{
\label{Fig.9.sub.4}
\includegraphics[width=0.32\textwidth]{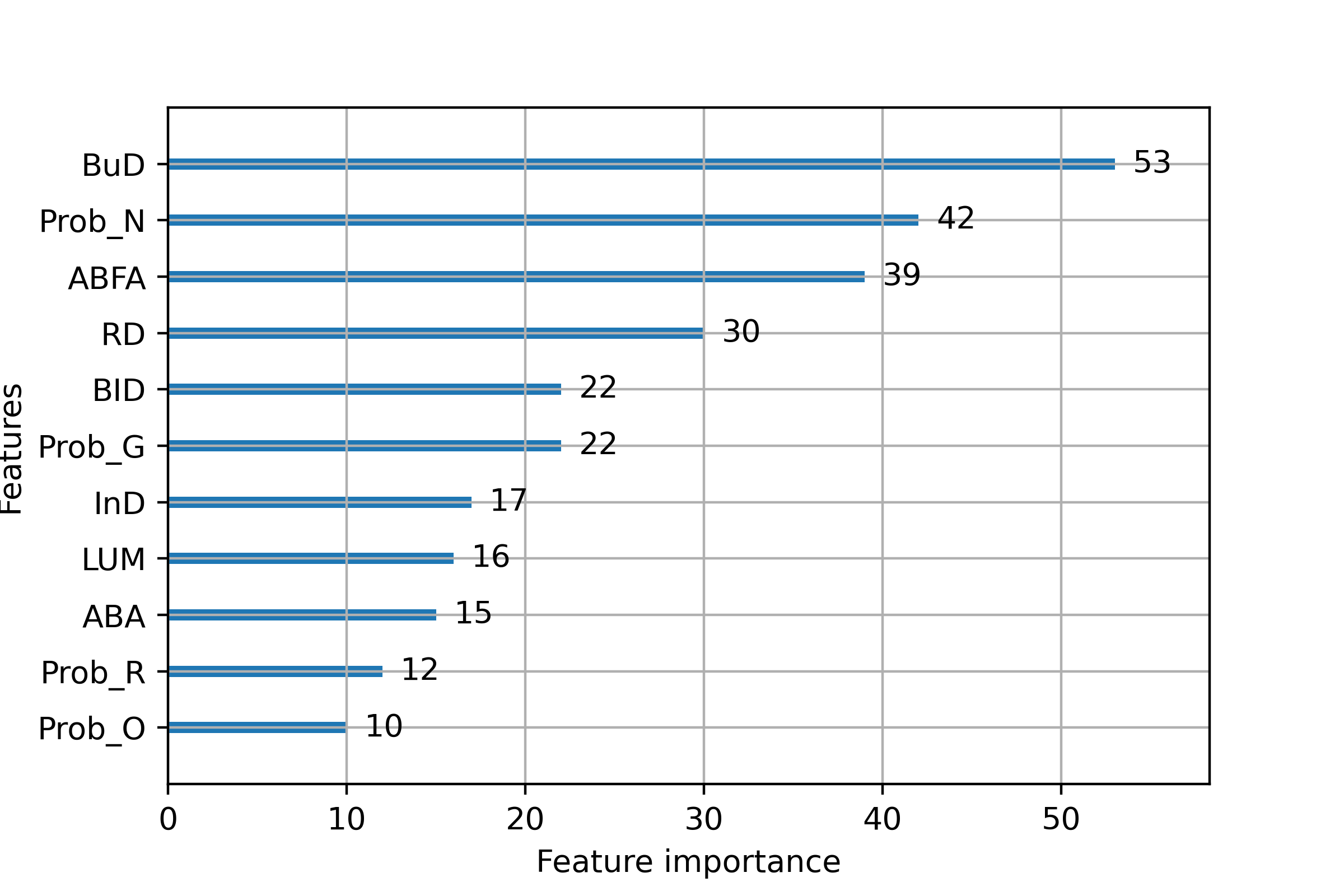}}
\subfigure[Airbnb density.]{
\label{Fig.9.sub.5}
\includegraphics[width=0.32\textwidth]{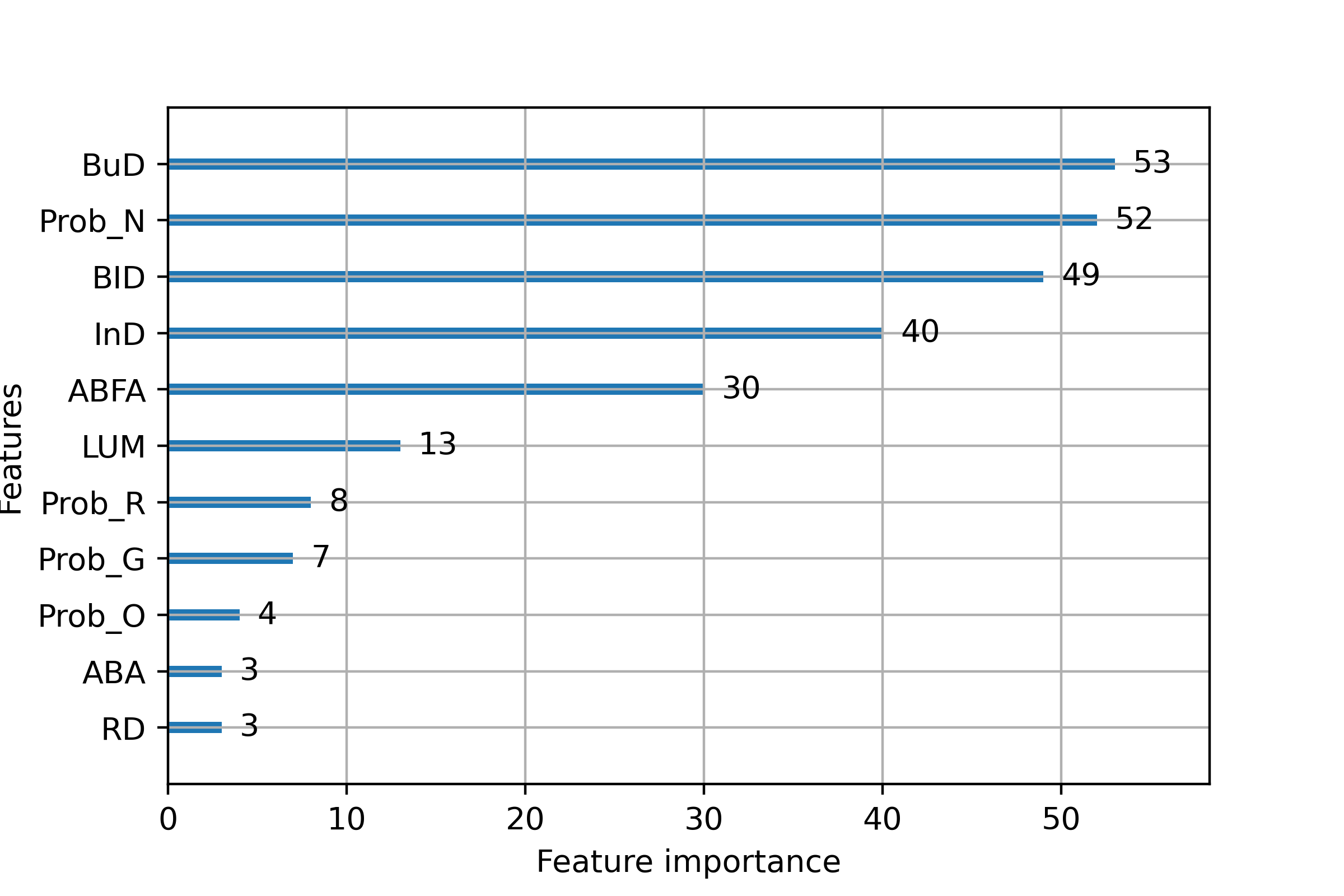}}
\subfigure[Vitality score.]{
\label{Fig.9.sub.0}
\includegraphics[width=0.32\textwidth]{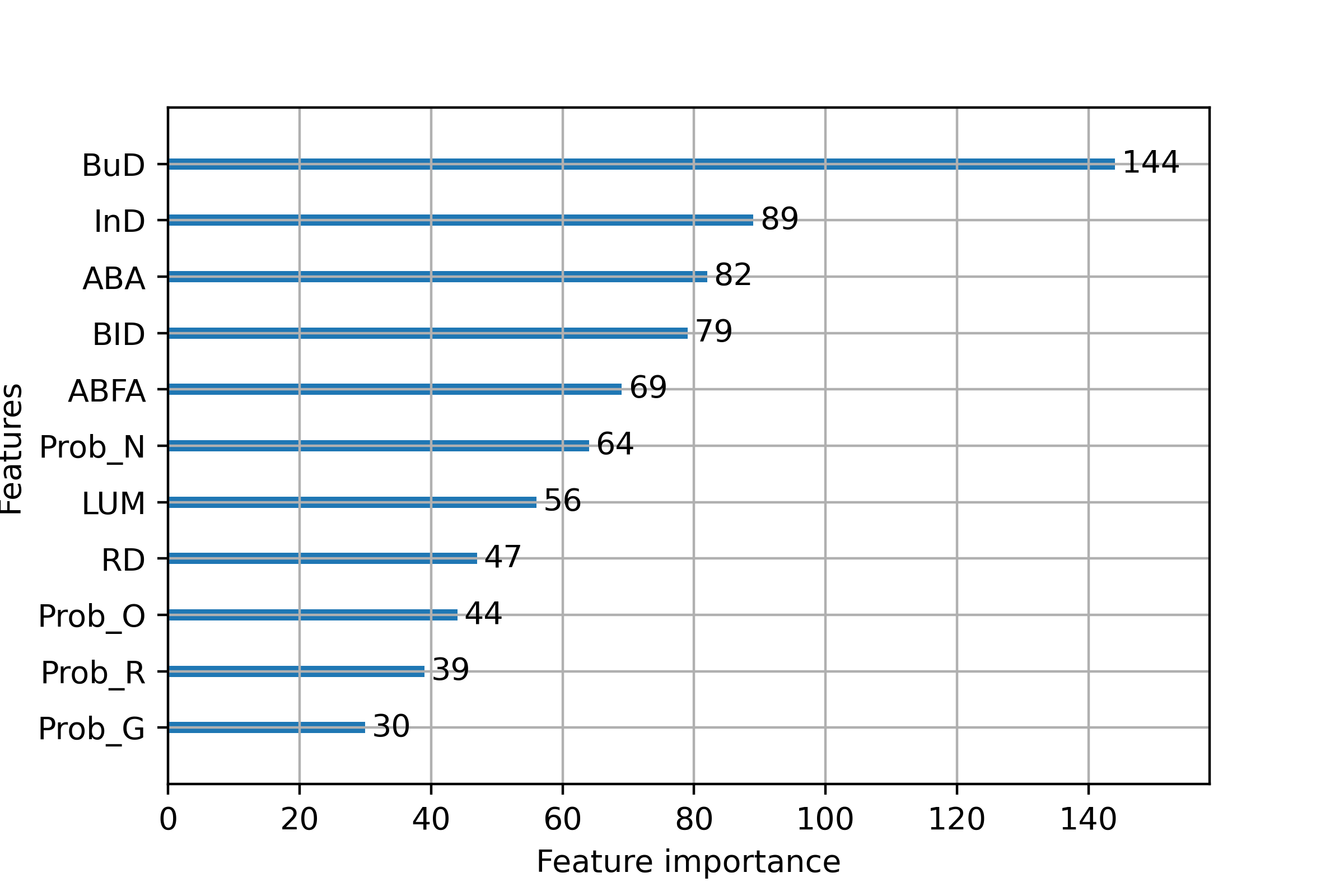}}
\caption{Feature importance plots of the overall model.}
\label{Fig.9.main}
\end{figure*}

While we have presented the differences of morphology-vitality relationship in regard to different subgroups of cities, this section focuses on the relationship between urban morphology and vitality in grid scale. Figure \ref{Fig.9.main} illustrates feature importance diagrams for the vitality score and the five vitality indicators of the corresponding models trained on the grids of all candidate cities. In this case, feature importance of a feature is calibrated by the number of times it is selected for splitting treenodes. As the total number of trees and treenodes vary from models, the absolute values of feature importance also differed from models but it would not affect our interpretation. The most important morphological features related to the POI density are building density, intersection density, and block density.  For tweet density, because of its near-zero R-squared, its importance feature could be misleading. The NTL turns out to be tied strongly with building density, average building footprint area and road density. Population and Airbnb density captures the strongest synergy with the probability of no pattern. Besides, building density, block density are also high ranking features for them. 

In conclusion, indicators regarding the development intensity (e.g. building density and intersection density) play a vital role in nearly all aspects of vitality, whereas probabilities of road network categories except for no pattern are in the opposite. Development intensity is the key for urban vitality in a grid, while comparatively patterned road network categories seem to matter not much. The feature importance diagram (Figure \ref{Fig.9.sub.0}) for the vitality score highlights analogous findings.

\section{Discussion}
\label{Sec.5}
\subsection{Understanding the effectiveness of road network classification}
\label{Sec.5.1}

\begin{table}
\centering
\small
\caption{Mean, median, and standard deviation of vitality scores of road network categories.}
\label{Tab.4}
\renewcommand\arraystretch{1.5}
\begin{tabular*}{8cm}{cccc}
\toprule
  &Mean& Median& Standard deviation\\
\midrule 
Gridiron&20.69&17.80&11.05\\
Organic&19.46&17.38&9.80\\
Radial&19.10&16.58&9.89\\
No pattern&10.86&9.91&5.12\\
\bottomrule
\end{tabular*}
\end{table}

\begin{figure*}[htbp]
\centering 
\subfigure[Kernel density estimation curves.]{
\label{Fig.10.sub.0}
\includegraphics[width=0.48\textwidth]{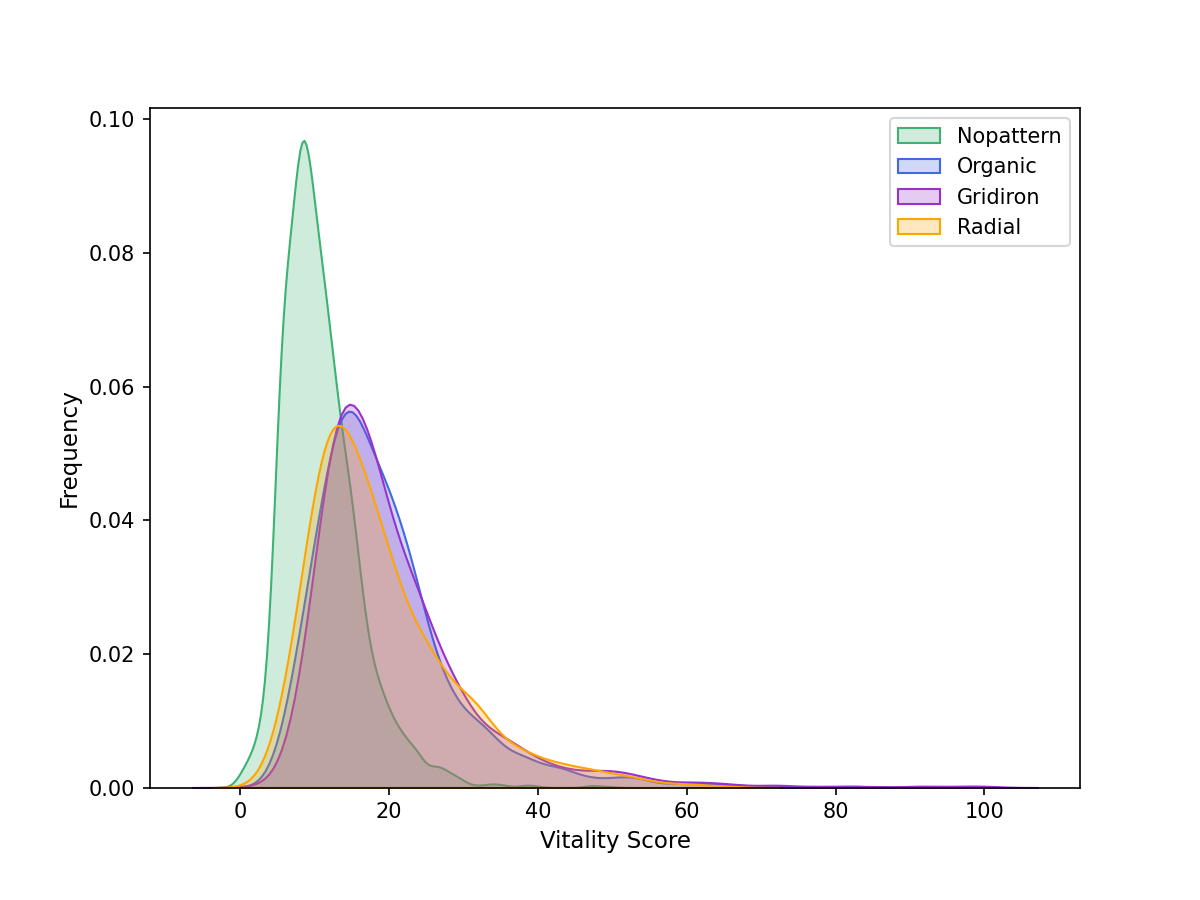}}
\subfigure[Lines: proportion of each road network category in ranges of vitality score. Bars: number of instances in the range.]{
\label{Fig.10.sub.1}
\includegraphics[width=0.48\textwidth]{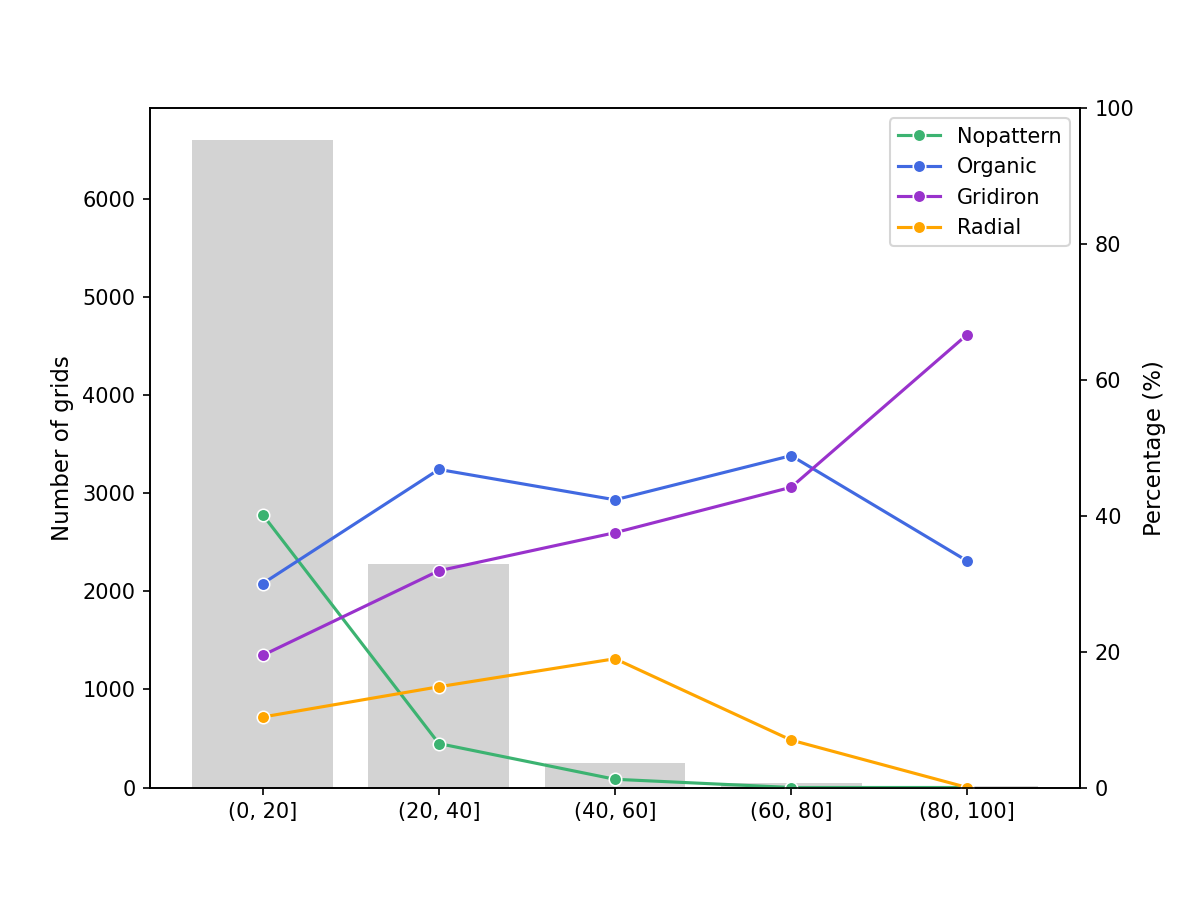}}
\caption{Distribution of vitality scores of four road network categories.}
\label{Fig.10.main}
\end{figure*}

Section \ref{sec.4.2} uncovers the small but positive effect of the road network classification. We would discuss and explore the root of it. Table \ref{Tab.4} lists the mean, median, and standard deviation of the overall vitality scores for each road network category. It underscores a clear gap of average vitality score between no pattern and the other categories. In contrast, the differences among the gridiron, organic, and radial grids are subtle. The kernel density estimation (KDE) curves shown in Figure \ref{Fig.10.sub.0} corroborate the finding. There is a clear left shift for the KDE curve of no pattern compared to the rests.
Given that salient correlations with the probability of no pattern exist for many aspects of vitality, it seems that the improvement brought by the classification primarily contributes to identifying urban areas without clear patterns.

To get a closer insight, we illustrate the proportion of each road network category in different ranges of vitality score together with the number of instances in each range in Figure \ref{Fig.10.sub.1}. From the first range (0-20) to the third range (40-60), the proportion of no pattern drops dramatically, making itself significant for the prediction. Meanwhile, the percentages of the other three categories grow in a similar rate, which means the differentiation among the three categories is trivial for the prediction. Given the fact that most instances concentrate on the first three ranges, it explains why, in global statistics, the probability of no pattern plays a role in urban vitality prediction while the rest three do not.

The no-patterns are those whose road networks fail to articulate a pattern or formulate one as efficient as the three patterned ones (gridiron, organic, and radial), which could root from manifold reasons. Subjective reasons could include locality in suburb, lying on land use requiring sparse network, etc. Objective reasons could be encountering geographic obstacle such as hills, lakes and parks. The area fitting the above description is relatively short of both human and activities. Also, the lack of well-structured networks causes less space for interaction and lower efficiency in all kinds of movements. All these result in lower level of vitality in no pattern area. Thus, probability of no pattern becomes helpful to urban vitality prediction.

\subsection{The influence of urban morphology on urban vitality}
\label{Sec.5.2}

According to the results and discussion presented earlier, globally, development intensity is predominant in determining urban vitality, and the identification of no pattern is the primary source of the effect of road network classification on urban vitality evaluation.

However, we also observe some anomalies in Figure \ref{Fig.10.sub.1}. We find that for the last two ranges with high vitality score (60-80 and 80-100), gridiron and organic covers an outright majority with the proportions of radial and no pattern approaching zero. The story deviates from the original trend that the patterned categories evenly distribute the proportion yielded from no pattern with the rising of vitality score. Instead, the relative proportion among the patterned categories becomes more unbalanced especially for the range 80-100. It means the probabilities of the patterned categories start to affect vitality evaluation in a way when vitality score is very high. Whereas, due to the very limited number of instances in these two ranges, the global statistics cannot capture and present this feature.

We further explore the particular case by listing the average vitality score and morphological indices of the top 10 vital instances for each patterned category (Table \ref{Tab.6}). In high vitality context, the gridirons surpass the counterparts much further in vitality score compared to the global case in Table \ref{Tab.4}. It should be noted that the radials keep the greatest building density whereas only achieve the lowest vitality score. In contrast, the gridirons and the organics keep lower building density but manage to maintain higher level of urban vitality. The finding indicates these two road network categories, especially gridiron may take advantage in the cultivation of extreme urban vitality. We attempt to discuss the reason behind this according to the information from Table \ref{Tab.6} and urban planner's intuition.

\begin{table*}
\centering
\small
\caption{Average morphological indices and vitality scores of top 10 vibrant grids for each patterned road network category.}
\label{Tab.6}
\renewcommand\arraystretch{1.5}
\begin{tabularx}{\textwidth}{lXXXXXXX}
\toprule
&Building density (\%)&Road density ($m/km^2$)&Intersection density ($km^{-2}$)&Average block area ($m^2$)&Block density ($km^{-2}$)&Average building footprint density ($m^2$) &Vitality score \\
\midrule 
Gridiron&58&22444.9&704.5&24484.37&126.3&1816.3&89.65\\
Organic&50&19801.5&675.6&8700.7&460.2&759.10&77.68\\
Radial&68&22212.4&584.8&40982.33&44.4&980.27&57.27\\
\bottomrule
\end{tabularx}
\end{table*}


In gridiron road networks, most buildings are surrounded by four orthogonal roads, so every facade of the buildings gets the opportunity to be interactive. Besides, the intensive and orthogonal road network bring about more intersections, as is shown in Table~\ref{Tab.6}. The increasing number of intersections is beneficial to the efficient movement of pedestrians and pave the way to incubate more human activities. All these advantages enable gridiron road network to bear extreme urban vitality.

As for the organics, they shared the characteristic of curved major roads that split the land into superblocks. Organic networks also expand the facades for interactions by increasing the length of the roads via curvature. It provides less efficient space for interactions than the gridirons, but requires less road density as well (Table~\ref{Tab.6}). It makes the organic a more economical option than the gridiron. The relatively stable high proportion of organic in all ranges supports this statement (Figure~\ref{Fig.10.sub.1}).

The radials have major roads from different directions converged at a central area, enabling the center to reach the surrounding areas efficiently and vice versa. The movement between blocks would more or less resort to the major network, which breeds more interactions on the facades facing major roads but in turn discourages the vibrancy of other areas. 
If the interactive facades of one radial structure are fully occupied, the spillover vitality resources (people, POIs, etc.) would prefer to mitigate to a better position elsewhere because of the comparative advantage. As a result, the vitality resources would be dispensed and results in the difficulty in achieving high level of vitality.

\subsection{Limitations and future studies}
The study proposed a new method to classify road networks and managed to fulfil a high level of classification accuracy. However, there is still space for the model to be ameliorated. First, the labelled image set still needs to be enriched with more samples around the world to make the model more universally applicable. Second, the model could recognize the overall pattern of the CRHDs well but not more detailed information. 
Object detection technique represented by Faster-RNN \citep{ren2016faster} and YOLO \citep{redmon2016look} could be an alternative solution. Instead of classifying the road network pattern with a fixed scale, object detection models may detect the road network patterns in variable scales within a larger CRHD, providing more information for planners.

This study included a sufficient number of distinct cities to assess the approach. Nevertheless, more cities should be involved in the future study for an all-round analysis. Conveniently, the process of generating a map of road network category such as the ones in Figure \ref{Fig.6.main} is automatic thanks to the CRHD generator and the road network classification model that we presented in this paper. It means that it would not be too complicated to add in more cities, which we seek to do in follow-up studies, potentially creating a global repository of urban form data. With that, perhaps we could obtain more conspicuous subgroups of road network categories. The difference between the subgroups, which was only briefly discussed in this paper due to length limitations, is also a topic worth digging deeper into.

A case study of urban vitality prediction was conducted in this paper to examine the effectiveness of road network classification, yet that may not have been perfect to confirm the effectiveness. Other case studies could also be made in the future with the supplement of the road classification model proposed in this paper. For instance, the influence of the road network category could be researched in other popular urban study fields such as energy consumption, heat island, mobility, and so forth. This new technique are expected to bring about more points of penetration in these research lines.

\section{Conclusions}
\label{Sec.6}

In this paper, we have presented a comprehensive and novel study investigating the nascent application of deep learning in urban morphology, and one that brings advancements also to research focusing on urban vitality.
We highlight the following five key contributions, findings, and takeaways of this study:

\begin{enumerate}
\item This research studied urban morphology from the perspective of road network classification. It proposed Colored Road Hierarchy Diagram (CRHD), which is easy to produce and beneficial for machine cognition. It offers an automatic pipeline to obtain the predicted category and corresponding probabilities for a road network, which offer several advantages over current approaches.
\item The study included the development of tools and models such as the Morphoindex generator, CHRD generator, and road network classification model, and we release them openly to facilitate future urban morphology studies.
\item Thanks to our approach, four subgroups were observed within the nine studied cities in terms of the percentage distribution of each sort of road networks. The approach is efficient and scalable, supporting large-scale comparative morphological studies at the global scale.
\item Aided by LightGBM, the effect of road network classification was examined through a case study of urban vitality prediction. The effect was small but positive. We detected the source of the effect. The measurement of vitality involved more aspects and data sources than previous studies.
\item Although development intensity was found the key factor of urban vitality globally, the study witnessed a possible role of road network categories in achieving high-level vitality. 
\end{enumerate}

The deep learning-based method to quantify urban morphology, which is the core of this paper, served as a road network classifier. It could output the probabilities of being gridiron, organic, radial, and no pattern for an inputted CRHD. The main body of the method was the road network classification model with the architecture of ResNet-34. It managed to attain an overall accuracy of 0.875. CRHD was proven to be more legible for machines. To some degree, we validated the potential capability of machines to perceive and understand urban morphology as humans do. The method we proposed showcases how to automate urban morphology study with supervised deep learning techniques.

We believe that the probabilities and categories obtained from the road classification model could help to reap more comprehensive knowledge on urban morphology and its impact on other urban study objects.
In our case study, we compared the performances between the baseline model and the augmented model in predicting overall urban vitality score.
We found that road network classification affects the vitality prediction positively but only to some degree. The association between morphological indices and each aspect of urban vitality was studied as well. Built environment vibrancy turned out to be the most relevant with urban morphology followed by population density and nighttime vibrancy. The cross-group differences in the morphology-vitality relationships was also discovered. Urban morphology of radial-organic-mixed cities is the most correlative with urban vitality. 

By analyzing the proportion of the road network categories in different ranges of vitality score, we confirmed the derivation of the detected effect of road network classification, which is the differentiation of no pattern area. We also found and discussed the possible disparity of ability to cultivate high level urban vitality among gridiron, organic and radial. Gridiron and organic seems more suitable for achieving high-level of urban vitality than radial. However, this result has not been fully proven in this study.

As a case study, the discussion on categorical urban morphology and urban vitality in this paper sheds a light on not only the improvement of statistic metrics but also the inspiration of thinking angles brought by our classification method. We hope to see the incorporation of our technique in other urban studies in the future.

\section*{Acknowledgments}
We gratefully acknowledge the data sources and the open-source packages used in this study.
We appreciate the comments by Jeffrey Ho (National University of Singapore), the members of the NUS Urban Analytics Lab, and the reviewers.
This research is part of the project Large-scale 3D Geospatial Data for Urban Analytics, which is supported by the National University of Singapore under the Start Up Grant R-295-000-171-133.



\bibliographystyle{elsarticle-harv}
\bibliography{reference}

\begin{thebibliography}{84}
\expandafter\ifx\csname natexlab\endcsname\relax\def\natexlab#1{#1}\fi
\providecommand{\url}[1]{\texttt{#1}}
\providecommand{\href}[2]{#2}
\providecommand{\path}[1]{#1}
\providecommand{\DOIprefix}{doi:}
\providecommand{\ArXivprefix}{arXiv:}
\providecommand{\URLprefix}{URL: }
\providecommand{\Pubmedprefix}{pmid:}
\providecommand{\doi}[1]{\href{http://dx.doi.org/#1}{\path{#1}}}
\providecommand{\Pubmed}[1]{\href{pmid:#1}{\path{#1}}}
\providecommand{\bibinfo}[2]{#2}
\ifx\xfnm\relax \def\xfnm[#1]{\unskip,\space#1}\fi
\bibitem[{Alexander(1977)}]{alexander1977pattern}
\bibinfo{author}{Alexander, C.}, \bibinfo{year}{1977}.
\newblock \bibinfo{title}{A pattern language: towns, buildings, construction}.
\newblock \bibinfo{publisher}{Oxford university press}.
\bibitem[{Baran et~al.(2008)Baran, Rodr{\'\i}guez and Khattak}]{baran2008space}
\bibinfo{author}{Baran, P.K.}, \bibinfo{author}{Rodr{\'\i}guez, D.A.},
  \bibinfo{author}{Khattak, A.J.}, \bibinfo{year}{2008}.
\newblock \bibinfo{title}{Space syntax and walking in a new urbanist and
  suburban neighbourhoods}.
\newblock \bibinfo{journal}{Journal of Urban Design} \bibinfo{volume}{13},
  \bibinfo{pages}{5--28}.
\bibitem[{Barrington-Leigh and Millard-Ball(2017)}]{BarringtonLeigh:2017ie}
\bibinfo{author}{Barrington-Leigh, C.}, \bibinfo{author}{Millard-Ball, A.},
  \bibinfo{year}{2017}.
\newblock \bibinfo{title}{{The world's user-generated road map is more than
  80\% complete}}.
\newblock \bibinfo{journal}{PLOS ONE} \bibinfo{volume}{12},
  \bibinfo{pages}{e0180698 -- 20}.
\newblock \DOIprefix\doi{10.1371/journal.pone.0180698}.
\bibitem[{Berghauser~Pont and Haupt(2007)}]{berghauser2007spacemate}
\bibinfo{author}{Berghauser~Pont, M.}, \bibinfo{author}{Haupt, P.},
  \bibinfo{year}{2007}.
\newblock \bibinfo{title}{The spacemate: Density and the typomorphology of the
  urban fabric}, in: \bibinfo{booktitle}{Urbanism laboratory for cities and
  regions: progress of research issues in urbanism}, pp.
  \bibinfo{pages}{11--26}.
\bibitem[{Berghauser-Pont and Haupt(2010)}]{berghauser2010spacematrix}
\bibinfo{author}{Berghauser-Pont, M.Y.}, \bibinfo{author}{Haupt, P.},
  \bibinfo{year}{2010}.
\newblock \bibinfo{title}{Spacematrix: space, density and urban form}.
\newblock \bibinfo{publisher}{NAi Publishers}, \bibinfo{address}{Rotterdam}.
\bibitem[{Biljecki(2020)}]{Biljecki.2020}
\bibinfo{author}{Biljecki, F.}, \bibinfo{year}{2020}.
\newblock \bibinfo{title}{{Exploration of open data in Southeast Asia to
  generate 3D building models}}.
\newblock \bibinfo{journal}{ISPRS Annals of Photogrammetry, Remote Sensing and
  Spatial Information Sciences} \bibinfo{volume}{VI-4/W1-2020},
  \bibinfo{pages}{37--44}.
\newblock \DOIprefix\doi{10.5194/isprs-annals-vi-4-w1-2020-37-2020}.
\bibitem[{Bocher et~al.(2018)Bocher, Petit, Bernard and
  Palominos}]{bocher2018geoprocessing}
\bibinfo{author}{Bocher, E.}, \bibinfo{author}{Petit, G.},
  \bibinfo{author}{Bernard, J.}, \bibinfo{author}{Palominos, S.},
  \bibinfo{year}{2018}.
\newblock \bibinfo{title}{A geoprocessing framework to compute urban
  indicators: The mapuce tools chain}.
\newblock \bibinfo{journal}{Urban climate} \bibinfo{volume}{24},
  \bibinfo{pages}{153--174}.
\bibitem[{Boeing(2017)}]{BOEING2017126}
\bibinfo{author}{Boeing, G.}, \bibinfo{year}{2017}.
\newblock \bibinfo{title}{Osmnx: New methods for acquiring, constructing,
  analyzing, and visualizing complex street networks}.
\newblock \bibinfo{journal}{Computers, Environment and Urban Systems}
  \bibinfo{volume}{65}, \bibinfo{pages}{126--139}.
\newblock \DOIprefix\doi{https://doi.org/10.1016/j.compenvurbsys.2017.05.004}.
\bibitem[{Boeing(2020)}]{boeing2020multi}
\bibinfo{author}{Boeing, G.}, \bibinfo{year}{2020}.
\newblock \bibinfo{title}{{A multi-scale analysis of 27,000 urban street
  networks: Every US city, town, urbanized area, and Zillow neighborhood}}.
\newblock \bibinfo{journal}{Environment and Planning B: Urban Analytics and
  City Science} \bibinfo{volume}{47}, \bibinfo{pages}{590--608}.
\bibitem[{Boeing(2021)}]{boeing2021spatial}
\bibinfo{author}{Boeing, G.}, \bibinfo{year}{2021}.
\newblock \bibinfo{title}{Spatial information and the legibility of urban form:
  Big data in urban morphology}.
\newblock \bibinfo{journal}{International Journal of Information Management}
  \bibinfo{volume}{56}, \bibinfo{pages}{102013}.
\bibitem[{Botta and Guti{\'e}rrez-Roig(2021)}]{botta2021modelling}
\bibinfo{author}{Botta, F.}, \bibinfo{author}{Guti{\'e}rrez-Roig, M.},
  \bibinfo{year}{2021}.
\newblock \bibinfo{title}{Modelling urban vibrancy with mobile phone and
  openstreetmap data}.
\newblock \bibinfo{journal}{Plos one} \bibinfo{volume}{16},
  \bibinfo{pages}{e0252015}.
\bibitem[{Canziani et~al.(2016)Canziani, Paszke and Culurciello}]{DBLP}
\bibinfo{author}{Canziani, A.}, \bibinfo{author}{Paszke, A.},
  \bibinfo{author}{Culurciello, E.}, \bibinfo{year}{2016}.
\newblock \bibinfo{title}{An analysis of deep neural network models for
  practical applications}.
\newblock \bibinfo{journal}{arXiv} \bibinfo{volume}{abs/1605.07678}.
\newblock \URLprefix \url{http://arxiv.org/abs/1605.07678},
  \href{http://arxiv.org/abs/1605.07678}{{\tt arXiv:1605.07678}}.
\bibitem[{Cao et~al.(2019)Cao, Guo and Zhang}]{cao2019comparison}
\bibinfo{author}{Cao, K.}, \bibinfo{author}{Guo, H.}, \bibinfo{author}{Zhang,
  Y.}, \bibinfo{year}{2019}.
\newblock \bibinfo{title}{{Comparison of approaches for urban functional zones
  classification based on multi-source geospatial data: a case study in Yuzhong
  District, Chongqing, China}}.
\newblock \bibinfo{journal}{Sustainability} \bibinfo{volume}{11},
  \bibinfo{pages}{660}.
\bibitem[{Chan et~al.(2011)Chan, Donner and L{\"a}mmer}]{chan2011urban}
\bibinfo{author}{Chan, S.H.}, \bibinfo{author}{Donner, R.V.},
  \bibinfo{author}{L{\"a}mmer, S.}, \bibinfo{year}{2011}.
\newblock \bibinfo{title}{Urban road networks—spatial networks with universal
  geometric features?}
\newblock \bibinfo{journal}{The European Physical Journal B}
  \bibinfo{volume}{84}, \bibinfo{pages}{563--577}.
\bibitem[{Chen et~al.(2019)Chen, Hui, Wu, Lang and Li}]{chen2019identifying}
\bibinfo{author}{Chen, T.}, \bibinfo{author}{Hui, E.C.}, \bibinfo{author}{Wu,
  J.}, \bibinfo{author}{Lang, W.}, \bibinfo{author}{Li, X.},
  \bibinfo{year}{2019}.
\newblock \bibinfo{title}{Identifying urban spatial structure and urban
  vibrancy in highly dense cities using georeferenced social media data}.
\newblock \bibinfo{journal}{Habitat International} \bibinfo{volume}{89},
  \bibinfo{pages}{102005}.
\bibitem[{Crooks et~al.(2016)Crooks, Croitoru, Jenkins, Mahabir, Agouris and
  Stefanidis}]{crooks2016user}
\bibinfo{author}{Crooks, A.}, \bibinfo{author}{Croitoru, A.},
  \bibinfo{author}{Jenkins, A.}, \bibinfo{author}{Mahabir, R.},
  \bibinfo{author}{Agouris, P.}, \bibinfo{author}{Stefanidis, A.},
  \bibinfo{year}{2016}.
\newblock \bibinfo{title}{User-generated big data and urban morphology}.
\newblock \bibinfo{journal}{Built Environment} \bibinfo{volume}{42},
  \bibinfo{pages}{396--414}.
\bibitem[{Delcl{\`o}s-Ali{\'o} and Miralles-Guasch(2018)}]{delclos2018looking}
\bibinfo{author}{Delcl{\`o}s-Ali{\'o}, X.}, \bibinfo{author}{Miralles-Guasch,
  C.}, \bibinfo{year}{2018}.
\newblock \bibinfo{title}{{Looking at Barcelona through Jane Jacobs’s eyes:
  Mapping the basic conditions for urban vitality in a Mediterranean
  conurbation}}.
\newblock \bibinfo{journal}{Land Use Policy} \bibinfo{volume}{75},
  \bibinfo{pages}{505--517}.
\bibitem[{Ding et~al.(2021)Ding, Fan and Gong}]{Ding.2021}
\bibinfo{author}{Ding, X.}, \bibinfo{author}{Fan, H.}, \bibinfo{author}{Gong,
  J.}, \bibinfo{year}{2021}.
\newblock \bibinfo{title}{{Towards generating network of bikeways from
  Mapillary data}}.
\newblock \bibinfo{journal}{Computers, Environment and Urban Systems}
  \bibinfo{volume}{88}, \bibinfo{pages}{101632}.
\newblock \DOIprefix\doi{10.1016/j.compenvurbsys.2021.101632}.
\bibitem[{Fleischmann(2019)}]{Fleischmann:2019fr}
\bibinfo{author}{Fleischmann, M.}, \bibinfo{year}{2019}.
\newblock \bibinfo{title}{{momepy: Urban Morphology Measuring Toolkit}}.
\newblock \bibinfo{journal}{Journal of Open Source Software}
  \bibinfo{volume}{4}, \bibinfo{pages}{1807 -- 4}.
\newblock \DOIprefix\doi{10.21105/joss.01807}.
\bibitem[{Fleischmann et~al.(2020)Fleischmann, Feliciotti, Romice and
  Porta}]{Fleischmann:2020fe}
\bibinfo{author}{Fleischmann, M.}, \bibinfo{author}{Feliciotti, A.},
  \bibinfo{author}{Romice, O.}, \bibinfo{author}{Porta, S.},
  \bibinfo{year}{2020}.
\newblock \bibinfo{title}{{Morphological tessellation as a way of partitioning
  space: Improving consistency in urban morphology at the plot scale}}.
\newblock \bibinfo{journal}{Computers, Environment and Urban Systems}
  \bibinfo{volume}{80}, \bibinfo{pages}{101441}.
\newblock \DOIprefix\doi{10.1016/j.compenvurbsys.2019.101441}.
\bibitem[{Garbasevschi et~al.(2021)Garbasevschi, Schmiedt, Verma, Lefter,
  Altes, Droin, Schiricke and Wurm}]{Garbasevschi.2021}
\bibinfo{author}{Garbasevschi, O.M.}, \bibinfo{author}{Schmiedt, J.E.},
  \bibinfo{author}{Verma, T.}, \bibinfo{author}{Lefter, I.},
  \bibinfo{author}{Altes, W.K.K.}, \bibinfo{author}{Droin, A.},
  \bibinfo{author}{Schiricke, B.}, \bibinfo{author}{Wurm, M.},
  \bibinfo{year}{2021}.
\newblock \bibinfo{title}{{Spatial factors influencing building age prediction
  and implications for urban residential energy modelling}}.
\newblock \bibinfo{journal}{Computers, Environment and Urban Systems}
  \bibinfo{volume}{88}, \bibinfo{pages}{101637}.
\newblock \DOIprefix\doi{10.1016/j.compenvurbsys.2021.101637}.
\bibitem[{Ge et~al.(2018)Ge, Yang, Zhu, Ma and Yang}]{ge2018ghost}
\bibinfo{author}{Ge, W.}, \bibinfo{author}{Yang, H.}, \bibinfo{author}{Zhu,
  X.}, \bibinfo{author}{Ma, M.}, \bibinfo{author}{Yang, Y.},
  \bibinfo{year}{2018}.
\newblock \bibinfo{title}{{Ghost city extraction and rate estimation in China
  based on npp-viirs night-time light data}}.
\newblock \bibinfo{journal}{ISPRS International Journal of Geo-Information}
  \bibinfo{volume}{7}, \bibinfo{pages}{219}.
\bibitem[{Gebru et~al.(2017)Gebru, Krause, Wang, Chen, Deng, Aiden and
  Fei-Fei}]{gebru2017using}
\bibinfo{author}{Gebru, T.}, \bibinfo{author}{Krause, J.},
  \bibinfo{author}{Wang, Y.}, \bibinfo{author}{Chen, D.},
  \bibinfo{author}{Deng, J.}, \bibinfo{author}{Aiden, E.L.},
  \bibinfo{author}{Fei-Fei, L.}, \bibinfo{year}{2017}.
\newblock \bibinfo{title}{{Using deep learning and Google Street View to
  estimate the demographic makeup of neighborhoods across the United States}}.
\newblock \bibinfo{journal}{Proceedings of the National Academy of Sciences}
  \bibinfo{volume}{114}, \bibinfo{pages}{13108--13113}.
\bibitem[{Han et~al.(2020)Han, Sun, Yu, Song and Ding}]{han2020classification}
\bibinfo{author}{Han, B.}, \bibinfo{author}{Sun, D.}, \bibinfo{author}{Yu, X.},
  \bibinfo{author}{Song, W.}, \bibinfo{author}{Ding, L.}, \bibinfo{year}{2020}.
\newblock \bibinfo{title}{Classification of urban street networks based on
  tree-like network features}.
\newblock \bibinfo{journal}{Sustainability} \bibinfo{volume}{12},
  \bibinfo{pages}{628}.
\bibitem[{He et~al.(2015)He, Zhang, Ren and Sun}]{he2015deep}
\bibinfo{author}{He, K.}, \bibinfo{author}{Zhang, X.}, \bibinfo{author}{Ren,
  S.}, \bibinfo{author}{Sun, J.}, \bibinfo{year}{2015}.
\newblock \bibinfo{title}{Deep residual learning for image recognition}.
\newblock \href{http://arxiv.org/abs/1512.03385}{{\tt arXiv:1512.03385}}.
\bibitem[{He et~al.(2016)He, Zhang, Ren and Sun}]{he2016identity}
\bibinfo{author}{He, K.}, \bibinfo{author}{Zhang, X.}, \bibinfo{author}{Ren,
  S.}, \bibinfo{author}{Sun, J.}, \bibinfo{year}{2016}.
\newblock \bibinfo{title}{Identity mappings in deep residual networks}.
\newblock \href{http://arxiv.org/abs/1603.05027}{{\tt arXiv:1603.05027}}.
\bibitem[{He et~al.(2018)He, He, Song, Wu, Yin and Mou}]{he2018impact}
\bibinfo{author}{He, Q.}, \bibinfo{author}{He, W.}, \bibinfo{author}{Song, Y.},
  \bibinfo{author}{Wu, J.}, \bibinfo{author}{Yin, C.}, \bibinfo{author}{Mou,
  Y.}, \bibinfo{year}{2018}.
\newblock \bibinfo{title}{The impact of urban growth patterns on urban vitality
  in newly built-up areas based on an association rules analysis using
  geographical `big data'}.
\newblock \bibinfo{journal}{Land Use Policy} \bibinfo{volume}{78},
  \bibinfo{pages}{726--738}.
\bibitem[{Helbich et~al.(2019)Helbich, Yao, Liu, Zhang, Liu and
  Wang}]{helbich2019using}
\bibinfo{author}{Helbich, M.}, \bibinfo{author}{Yao, Y.}, \bibinfo{author}{Liu,
  Y.}, \bibinfo{author}{Zhang, J.}, \bibinfo{author}{Liu, P.},
  \bibinfo{author}{Wang, R.}, \bibinfo{year}{2019}.
\newblock \bibinfo{title}{{Using deep learning to examine street view green and
  blue spaces and their associations with geriatric depression in Beijing,
  China}}.
\newblock \bibinfo{journal}{Environment international} \bibinfo{volume}{126},
  \bibinfo{pages}{107--117}.
\bibitem[{Hillier(1996)}]{hillier1996space}
\bibinfo{author}{Hillier, B.}, \bibinfo{year}{1996}.
\newblock \bibinfo{title}{Space is the machine: a configurational theory of
  architecture}.
\newblock \bibinfo{publisher}{Cambridge University Press},
  \bibinfo{address}{Cambridge}.
\bibitem[{Hillier and Hanson(1984;1989;1988;)}]{hillier1984social}
\bibinfo{author}{Hillier, B.}, \bibinfo{author}{Hanson, J.},
  \bibinfo{year}{1984;1989;1988;}.
\newblock \bibinfo{title}{The social logic of space}.
\newblock \bibinfo{publisher}{Cambridge University Press},
  \bibinfo{address}{Cambridge}.
\bibitem[{Jacobs(1961)}]{jacobs1961death}
\bibinfo{author}{Jacobs, J.}, \bibinfo{year}{1961}.
\newblock \bibinfo{title}{{The Death and Life of Great American Cities}}.
\newblock \bibinfo{publisher}{Randoms House, New York}.
\bibitem[{Jiang and Claramunt(2002)}]{jiang2002integration}
\bibinfo{author}{Jiang, B.}, \bibinfo{author}{Claramunt, C.},
  \bibinfo{year}{2002}.
\newblock \bibinfo{title}{{Integration of space syntax into GIS: new
  perspectives for urban morphology}}.
\newblock \bibinfo{journal}{Transactions in GIS} \bibinfo{volume}{6},
  \bibinfo{pages}{295--309}.
\bibitem[{Jin et~al.(2017)Jin, Long, Sun, Lu, Yang and
  Tang}]{jin2017evaluating}
\bibinfo{author}{Jin, X.}, \bibinfo{author}{Long, Y.}, \bibinfo{author}{Sun,
  W.}, \bibinfo{author}{Lu, Y.}, \bibinfo{author}{Yang, X.},
  \bibinfo{author}{Tang, J.}, \bibinfo{year}{2017}.
\newblock \bibinfo{title}{{Evaluating cities' vitality and identifying ghost
  cities in China with emerging geographical data}}.
\newblock \bibinfo{journal}{Cities} \bibinfo{volume}{63},
  \bibinfo{pages}{98--109}.
\bibitem[{Jochem and Tatem(2021)}]{Jochem.2021}
\bibinfo{author}{Jochem, W.C.}, \bibinfo{author}{Tatem, A.J.},
  \bibinfo{year}{2021}.
\newblock \bibinfo{title}{{Tools for mapping multi-scale settlement patterns of
  building footprints: An introduction to the R package foot}}.
\newblock \bibinfo{journal}{PLOS ONE} \bibinfo{volume}{16},
  \bibinfo{pages}{e0247535}.
\newblock \DOIprefix\doi{10.1371/journal.pone.0247535}.
\bibitem[{Ke et~al.(2017)Ke, Meng, Finley, Wang, Chen, Ma, Ye and
  Liu}]{ke2017lightgbm}
\bibinfo{author}{Ke, G.}, \bibinfo{author}{Meng, Q.}, \bibinfo{author}{Finley,
  T.}, \bibinfo{author}{Wang, T.}, \bibinfo{author}{Chen, W.},
  \bibinfo{author}{Ma, W.}, \bibinfo{author}{Ye, Q.}, \bibinfo{author}{Liu,
  T.Y.}, \bibinfo{year}{2017}.
\newblock \bibinfo{title}{Lightgbm: A highly efficient gradient boosting
  decision tree}.
\newblock \bibinfo{journal}{Advances in neural information processing systems}
  \bibinfo{volume}{30}, \bibinfo{pages}{3146--3154}.
\bibitem[{Kim et~al.(2021)Kim, Lee, Hipp and Ki}]{kim2021decoding}
\bibinfo{author}{Kim, J.H.}, \bibinfo{author}{Lee, S.}, \bibinfo{author}{Hipp,
  J.R.}, \bibinfo{author}{Ki, D.}, \bibinfo{year}{2021}.
\newblock \bibinfo{title}{Decoding urban landscapes: Google street view and
  measurement sensitivity}.
\newblock \bibinfo{journal}{Computers, Environment and Urban Systems}
  \bibinfo{volume}{88}, \bibinfo{pages}{101626}.
\bibitem[{Kim(2018)}]{kim2018seoul}
\bibinfo{author}{Kim, Y.L.}, \bibinfo{year}{2018}.
\newblock \bibinfo{title}{Seoul's wi-fi hotspots: Wi-fi access points as an
  indicator of urban vitality}.
\newblock \bibinfo{journal}{Computers, Environment and Urban Systems}
  \bibinfo{volume}{72}, \bibinfo{pages}{13--24}.
\bibitem[{Kim(2020)}]{kim2020data}
\bibinfo{author}{Kim, Y.L.}, \bibinfo{year}{2020}.
\newblock \bibinfo{title}{{Data-driven approach to characterize urban vitality:
  how spatiotemporal context dynamically defines Seoul's nighttime}}.
\newblock \bibinfo{journal}{International Journal of Geographical Information
  Science} \bibinfo{volume}{34}, \bibinfo{pages}{1235--1256}.
\bibitem[{Landry(2000)}]{landry2000urbann}
\bibinfo{author}{Landry, C.}, \bibinfo{year}{2000}.
\newblock \bibinfo{title}{Urban vitality: A new source of urban
  competitiveness}.
\newblock \bibinfo{journal}{Archis} , \bibinfo{pages}{8--13}.
\bibitem[{Li and Biljecki(2019)}]{Li:2019vv}
\bibinfo{author}{Li, J.}, \bibinfo{author}{Biljecki, F.}, \bibinfo{year}{2019}.
\newblock \bibinfo{title}{{The implementation of big data analysis in
  regulating online short-term rental business: a case of Airbnb in Beijing}}.
\newblock \bibinfo{journal}{ISPRS Ann. Photogramm. Remote Sens. Spatial Inf.
  Sci.} \bibinfo{volume}{IV-4/W9}, \bibinfo{pages}{79 -- 86}.
\newblock \DOIprefix\doi{10.5194/isprs-annals-iv-4-w9-79-2019}.
\bibitem[{Li et~al.(2020a)Li, Wu, Lin, Li and Du}]{li2020urban}
\bibinfo{author}{Li, S.}, \bibinfo{author}{Wu, C.}, \bibinfo{author}{Lin, Y.},
  \bibinfo{author}{Li, Z.}, \bibinfo{author}{Du, Q.}, \bibinfo{year}{2020}a.
\newblock \bibinfo{title}{{Urban Morphology Promotes Urban Vibrancy from the
  Spatiotemporal and Synergetic Perspectives: A Case Study Using Multisource
  Data in Shenzhen, China}}.
\newblock \bibinfo{journal}{Sustainability} \bibinfo{volume}{12},
  \bibinfo{pages}{4829}.
\bibitem[{Li et~al.(2020b)Li, Cheng, Lv, Song, Jia and Lu}]{li2020data}
\bibinfo{author}{Li, X.}, \bibinfo{author}{Cheng, S.}, \bibinfo{author}{Lv,
  Z.}, \bibinfo{author}{Song, H.}, \bibinfo{author}{Jia, T.},
  \bibinfo{author}{Lu, N.}, \bibinfo{year}{2020}b.
\newblock \bibinfo{title}{Data analytics of urban fabric metrics for smart
  cities}.
\newblock \bibinfo{journal}{Future Generation Computer Systems}
  \bibinfo{volume}{107}, \bibinfo{pages}{871--882}.
\bibitem[{Liu et~al.(2010)Liu, Xu, Jiang and Wu}]{liu2010evaluation}
\bibinfo{author}{Liu, L.}, \bibinfo{author}{Xu, Y.l.}, \bibinfo{author}{Jiang,
  S.}, \bibinfo{author}{Wu, Q.}, \bibinfo{year}{2010}.
\newblock \bibinfo{title}{Evaluation of urban vitality based on fuzzy
  matter-element mode}.
\newblock \bibinfo{journal}{Geography and Geo-Information Science}
  \bibinfo{volume}{26}, \bibinfo{pages}{73--77}.
\bibitem[{Lloyd et~al.(2019)Lloyd, Chamberlain, Kerr, Yetman, Pistolesi,
  Stevens, Gaughan, Nieves, Hornby, MacManus, Sinha, Bondarenko, Sorichetta and
  Tatem}]{Lloyd:2019gc}
\bibinfo{author}{Lloyd, C.T.}, \bibinfo{author}{Chamberlain, H.},
  \bibinfo{author}{Kerr, D.}, \bibinfo{author}{Yetman, G.},
  \bibinfo{author}{Pistolesi, L.}, \bibinfo{author}{Stevens, F.R.},
  \bibinfo{author}{Gaughan, A.E.}, \bibinfo{author}{Nieves, J.J.},
  \bibinfo{author}{Hornby, G.}, \bibinfo{author}{MacManus, K.},
  \bibinfo{author}{Sinha, P.}, \bibinfo{author}{Bondarenko, M.},
  \bibinfo{author}{Sorichetta, A.}, \bibinfo{author}{Tatem, A.J.},
  \bibinfo{year}{2019}.
\newblock \bibinfo{title}{{Global spatio-temporally harmonised datasets for
  producing high-resolution gridded population distribution datasets}}.
\newblock \bibinfo{journal}{Big Earth Data} \bibinfo{volume}{3},
  \bibinfo{pages}{108 -- 139}.
\newblock \DOIprefix\doi{10.1080/20964471.2019.1625151}.
\bibitem[{Lloyd et~al.(2017)Lloyd, Sorichetta and Tatem}]{Lloyd:2017et}
\bibinfo{author}{Lloyd, C.T.}, \bibinfo{author}{Sorichetta, A.},
  \bibinfo{author}{Tatem, A.J.}, \bibinfo{year}{2017}.
\newblock \bibinfo{title}{{High resolution global gridded data for use in
  population studies}}.
\newblock \bibinfo{journal}{Scientific Data} \bibinfo{volume}{4},
  \bibinfo{pages}{e1298}.
\newblock \DOIprefix\doi{10.1038/sdata.2017.1}.
\bibitem[{Lopes and Camanho(2013)}]{lopes2013public}
\bibinfo{author}{Lopes, M.N.}, \bibinfo{author}{Camanho, A.S.},
  \bibinfo{year}{2013}.
\newblock \bibinfo{title}{Public green space use and consequences on urban
  vitality: An assessment of european cities}.
\newblock \bibinfo{journal}{Social indicators research} \bibinfo{volume}{113},
  \bibinfo{pages}{751--767}.
\bibitem[{Lynch(1984)}]{lynch1984good}
\bibinfo{author}{Lynch, K.}, \bibinfo{year}{1984}.
\newblock \bibinfo{title}{Good city form}.
\newblock \bibinfo{publisher}{MIT press}.
\bibitem[{Ma et~al.(2014)Ma, Zhou, Wang, Zhou, Haynie and Xu}]{ma2014diverse}
\bibinfo{author}{Ma, T.}, \bibinfo{author}{Zhou, Y.}, \bibinfo{author}{Wang,
  Y.}, \bibinfo{author}{Zhou, C.}, \bibinfo{author}{Haynie, S.},
  \bibinfo{author}{Xu, T.}, \bibinfo{year}{2014}.
\newblock \bibinfo{title}{Diverse relationships between suomi-npp viirs
  night-time light and multi-scale socioeconomic activity}.
\newblock \bibinfo{journal}{Remote Sensing Letters} \bibinfo{volume}{5},
  \bibinfo{pages}{652--661}.
\bibitem[{Marshall and Gong(2005)}]{marshall2005urban}
\bibinfo{author}{Marshall, S.}, \bibinfo{author}{Gong, Y.},
  \bibinfo{year}{2005}.
\newblock \bibinfo{title}{Urban pattern specification}.
\newblock \bibinfo{publisher}{Institute of Community Studies, London}.
\bibitem[{Martino et~al.(2021)Martino, Girling and Lu}]{martino2021urban}
\bibinfo{author}{Martino, N.}, \bibinfo{author}{Girling, C.},
  \bibinfo{author}{Lu, Y.}, \bibinfo{year}{2021}.
\newblock \bibinfo{title}{Urban form and livability: socioeconomic and built
  environment indicators}.
\newblock \bibinfo{journal}{Buildings and Cities} \bibinfo{volume}{2}.
\bibitem[{Meng and Xing(2019)}]{meng2019exploring}
\bibinfo{author}{Meng, Y.}, \bibinfo{author}{Xing, H.}, \bibinfo{year}{2019}.
\newblock \bibinfo{title}{Exploring the relationship between landscape
  characteristics and urban vibrancy: A case study using morphology and review
  data}.
\newblock \bibinfo{journal}{Cities} \bibinfo{volume}{95},
  \bibinfo{pages}{102389}.
\bibitem[{Middel et~al.(2019)Middel, Lukasczyk, Zakrzewski, Arnold and
  Maciejewski}]{middel2019urban}
\bibinfo{author}{Middel, A.}, \bibinfo{author}{Lukasczyk, J.},
  \bibinfo{author}{Zakrzewski, S.}, \bibinfo{author}{Arnold, M.},
  \bibinfo{author}{Maciejewski, R.}, \bibinfo{year}{2019}.
\newblock \bibinfo{title}{Urban form and composition of street canyons: A
  human-centric big data and deep learning approach}.
\newblock \bibinfo{journal}{Landscape and Urban Planning}
  \bibinfo{volume}{183}, \bibinfo{pages}{122--132}.
\bibitem[{Moosavi(2017)}]{moosavi2017urban}
\bibinfo{author}{Moosavi, V.}, \bibinfo{year}{2017}.
\newblock \bibinfo{title}{Urban morphology meets deep learning: Exploring urban
  forms in one million cities, town and villages across the planet}.
\newblock \bibinfo{journal}{arXiv preprint arXiv:1709.02939} .
\bibitem[{Moudon(1997)}]{moudon1997urban}
\bibinfo{author}{Moudon, A.V.}, \bibinfo{year}{1997}.
\newblock \bibinfo{title}{Urban morphology as an emerging interdisciplinary
  field}.
\newblock \bibinfo{journal}{Urban morphology} \bibinfo{volume}{1},
  \bibinfo{pages}{3--10}.
\bibitem[{Plater-Zyberk et~al.(2003)Plater-Zyberk, Longo, Hetzel, Davis and
  Duany}]{plater2003lexicon}
\bibinfo{author}{Plater-Zyberk, E.}, \bibinfo{author}{Longo, G.},
  \bibinfo{author}{Hetzel, P.}, \bibinfo{author}{Davis, R.},
  \bibinfo{author}{Duany, A.}, \bibinfo{year}{2003}.
\newblock \bibinfo{title}{The lexicon of new urbanism}.
\newblock \bibinfo{journal}{Duany Plater-Zyberk \& Company} .
\bibitem[{Qu et~al.(2019)Qu, Leng and Ma}]{qu2019investigating}
\bibinfo{author}{Qu, B.}, \bibinfo{author}{Leng, J.}, \bibinfo{author}{Ma, J.},
  \bibinfo{year}{2019}.
\newblock \bibinfo{title}{{Investigating the Intensive Redevelopment of Urban
  Central Blocks Using Data Envelopment Analysis and Deep Learning: A Case
  Study of Nanjing, China}}.
\newblock \bibinfo{journal}{IEEE Access} \bibinfo{volume}{7},
  \bibinfo{pages}{109884--109898}.
\bibitem[{Redmon et~al.(2016)Redmon, Divvala, Girshick and
  Farhadi}]{redmon2016look}
\bibinfo{author}{Redmon, J.}, \bibinfo{author}{Divvala, S.},
  \bibinfo{author}{Girshick, R.}, \bibinfo{author}{Farhadi, A.},
  \bibinfo{year}{2016}.
\newblock \bibinfo{title}{You only look once: Unified, real-time object
  detection}.
\newblock \href{http://arxiv.org/abs/1506.02640}{{\tt arXiv:1506.02640}}.
\bibitem[{Ren et~al.(2016)Ren, He, Girshick and Sun}]{ren2016faster}
\bibinfo{author}{Ren, S.}, \bibinfo{author}{He, K.}, \bibinfo{author}{Girshick,
  R.}, \bibinfo{author}{Sun, J.}, \bibinfo{year}{2016}.
\newblock \bibinfo{title}{{Faster R-CNN: Towards Real-Time Object Detection
  with Region Proposal Networks}}.
\newblock \href{http://arxiv.org/abs/1506.01497}{{\tt arXiv:1506.01497}}.
\bibitem[{Shannon(1948)}]{shannon1948mathematical}
\bibinfo{author}{Shannon, C.E.}, \bibinfo{year}{1948}.
\newblock \bibinfo{title}{A mathematical theory of communication}.
\newblock \bibinfo{journal}{The Bell system technical journal}
  \bibinfo{volume}{27}, \bibinfo{pages}{379--423}.
\bibitem[{Snellen et~al.(2002)Snellen, Borgers and
  Timmermans}]{Danielle2002urban}
\bibinfo{author}{Snellen, D.}, \bibinfo{author}{Borgers, A.},
  \bibinfo{author}{Timmermans, H.}, \bibinfo{year}{2002}.
\newblock \bibinfo{title}{Urban form, road network type, and mode choice for
  frequently conducted activities: A multilevel analysis using
  quasi-experimental design data}.
\newblock \bibinfo{journal}{Environment and Planning A: Economy and Space}
  \bibinfo{volume}{34}, \bibinfo{pages}{1207--1220}.
\newblock \URLprefix \url{https://doi.org/10.1068/a349},
  \DOIprefix\doi{10.1068/a349}.
\bibitem[{Southworth and Ben-Joseph(1995)}]{southworth1995street}
\bibinfo{author}{Southworth, M.}, \bibinfo{author}{Ben-Joseph, E.},
  \bibinfo{year}{1995}.
\newblock \bibinfo{title}{Street standards and the shaping of suburbia}.
\newblock \bibinfo{journal}{Journal of the American Planning Association}
  \bibinfo{volume}{61}, \bibinfo{pages}{65--81}.
\bibitem[{Southworth and Ben-Joseph(2013)}]{southworth2013streets}
\bibinfo{author}{Southworth, M.}, \bibinfo{author}{Ben-Joseph, E.},
  \bibinfo{year}{2013}.
\newblock \bibinfo{title}{Streets and the Shaping of Towns and Cities}.
\newblock \bibinfo{publisher}{Island Press}.
\bibitem[{Sung and Lee(2015)}]{sung2015residential}
\bibinfo{author}{Sung, H.}, \bibinfo{author}{Lee, S.}, \bibinfo{year}{2015}.
\newblock \bibinfo{title}{{Residential built environment and walking activity:
  Empirical evidence of Jane Jacobs' urban vitality}}.
\newblock \bibinfo{journal}{Transportation Research Part D: Transport and
  Environment} \bibinfo{volume}{41}, \bibinfo{pages}{318--329}.
\bibitem[{Tatem(2017)}]{Tatem.2017}
\bibinfo{author}{Tatem, A.J.}, \bibinfo{year}{2017}.
\newblock \bibinfo{title}{{WorldPop, open data for spatial demography}}.
\newblock \bibinfo{journal}{Scientific Data} \bibinfo{volume}{4},
  \bibinfo{pages}{170004}.
\newblock \DOIprefix\doi{10.1038/sdata.2017.4}.
\bibitem[{Wang and Vermeulen(2020)}]{wang2020life}
\bibinfo{author}{Wang, M.}, \bibinfo{author}{Vermeulen, F.},
  \bibinfo{year}{2020}.
\newblock \bibinfo{title}{Life between buildings from a street view image: What
  do big data analytics reveal about neighbourhood organisational vitality?}
\newblock \bibinfo{journal}{Urban Studies} , \bibinfo{pages}{0042098020957198}.
\bibitem[{WorldPop(2018)}]{worldpop}
\bibinfo{author}{WorldPop}, \bibinfo{year}{2018}.
\newblock \bibinfo{title}{Global 1km population}.
\newblock \DOIprefix\doi{10.5258/SOTON/WP00647}.
\bibitem[{Wu and Biljecki(2021)}]{Wu.2020pdi}
\bibinfo{author}{Wu, A.N.}, \bibinfo{author}{Biljecki, F.},
  \bibinfo{year}{2021}.
\newblock \bibinfo{title}{{Roofpedia: Automatic mapping of green and solar
  roofs for an open roofscape registry and evaluation of urban
  sustainability}}.
\newblock \bibinfo{journal}{Landscape and Urban Planning}
  \bibinfo{volume}{214}, \bibinfo{pages}{104167}.
\newblock \DOIprefix\doi{10.1016/j.landurbplan.2021.104167}.
\bibitem[{Wu et~al.(2018)Wu, Ye, Ren and Du}]{wu2018check}
\bibinfo{author}{Wu, C.}, \bibinfo{author}{Ye, X.}, \bibinfo{author}{Ren, F.},
  \bibinfo{author}{Du, Q.}, \bibinfo{year}{2018}.
\newblock \bibinfo{title}{{Check-in behaviour and spatio-temporal vibrancy: An
  exploratory analysis in Shenzhen, China}}.
\newblock \bibinfo{journal}{Cities} \bibinfo{volume}{77},
  \bibinfo{pages}{104--116}.
\bibitem[{Wu and Niu(2019)}]{wu2019influence}
\bibinfo{author}{Wu, W.}, \bibinfo{author}{Niu, X.}, \bibinfo{year}{2019}.
\newblock \bibinfo{title}{Influence of built environment on urban vitality:
  Case study of shanghai using mobile phone location data}.
\newblock \bibinfo{journal}{Journal of Urban Planning and Development}
  \bibinfo{volume}{145}, \bibinfo{pages}{04019007}.
\bibitem[{Xia et~al.(2020)Xia, Yeh and Zhang}]{xia2020analyzing}
\bibinfo{author}{Xia, C.}, \bibinfo{author}{Yeh, A.G.O.},
  \bibinfo{author}{Zhang, A.}, \bibinfo{year}{2020}.
\newblock \bibinfo{title}{{Analyzing spatial relationships between urban land
  use intensity and urban vitality at street block level: A case study of five
  Chinese megacities}}.
\newblock \bibinfo{journal}{Landscape and Urban Planning}
  \bibinfo{volume}{193}, \bibinfo{pages}{103669}.
\bibitem[{Xiao et~al.(2017)Xiao, Wang, Fu and Wu}]{xiao2017identifying}
\bibinfo{author}{Xiao, Z.}, \bibinfo{author}{Wang, Y.}, \bibinfo{author}{Fu,
  K.}, \bibinfo{author}{Wu, F.}, \bibinfo{year}{2017}.
\newblock \bibinfo{title}{Identifying different transportation modes from
  trajectory data using tree-based ensemble classifiers}.
\newblock \bibinfo{journal}{ISPRS International Journal of Geo-Information}
  \bibinfo{volume}{6}, \bibinfo{pages}{57}.
\bibitem[{Yang et~al.(2021)Yang, Cao and Zhou}]{yang2021elaborating}
\bibinfo{author}{Yang, J.}, \bibinfo{author}{Cao, J.}, \bibinfo{author}{Zhou,
  Y.}, \bibinfo{year}{2021}.
\newblock \bibinfo{title}{{Elaborating non-linear associations and synergies of
  subway access and land uses with urban vitality in Shenzhen}}.
\newblock \bibinfo{journal}{Transportation Research Part A: Policy and
  Practice} \bibinfo{volume}{144}, \bibinfo{pages}{74--88}.
\bibitem[{Ye et~al.(2018)Ye, Li and Liu}]{ye2018block}
\bibinfo{author}{Ye, Y.}, \bibinfo{author}{Li, D.}, \bibinfo{author}{Liu, X.},
  \bibinfo{year}{2018}.
\newblock \bibinfo{title}{{How block density and typology affect urban
  vitality: An exploratory analysis in Shenzhen, China}}.
\newblock \bibinfo{journal}{Urban Geography} \bibinfo{volume}{39},
  \bibinfo{pages}{631--652}.
\bibitem[{Ye and Van~Nes(2014)}]{ye2014quantitative}
\bibinfo{author}{Ye, Y.}, \bibinfo{author}{Van~Nes, A.}, \bibinfo{year}{2014}.
\newblock \bibinfo{title}{Quantitative tools in urban morphology: Combining
  space syntax, spacematrix and mixed-use index in a gis framework}.
\newblock \bibinfo{journal}{Urban morphology} \bibinfo{volume}{18},
  \bibinfo{pages}{97--118}.
\bibitem[{Yuan et~al.(2019)Yuan, Shan, Zhang, Li, Yin, Hang and
  Norford}]{Yuan.2019}
\bibinfo{author}{Yuan, C.}, \bibinfo{author}{Shan, R.}, \bibinfo{author}{Zhang,
  Y.}, \bibinfo{author}{Li, X.X.}, \bibinfo{author}{Yin, T.},
  \bibinfo{author}{Hang, J.}, \bibinfo{author}{Norford, L.},
  \bibinfo{year}{2019}.
\newblock \bibinfo{title}{{Multilayer urban canopy modelling and mapping for
  traffic pollutant dispersion at high density urban areas}}.
\newblock \bibinfo{journal}{Science of The Total Environment}
  \bibinfo{volume}{647}, \bibinfo{pages}{255--267}.
\newblock \DOIprefix\doi{10.1016/j.scitotenv.2018.07.409}.
\bibitem[{Yue et~al.(2019)Yue, Chen, Zhang and Liu}]{yue2019spatial}
\bibinfo{author}{Yue, W.}, \bibinfo{author}{Chen, Y.}, \bibinfo{author}{Zhang,
  Q.}, \bibinfo{author}{Liu, Y.}, \bibinfo{year}{2019}.
\newblock \bibinfo{title}{{Spatial explicit assessment of urban vitality using
  multi-source data: A case of Shanghai, China}}.
\newblock \bibinfo{journal}{Sustainability} \bibinfo{volume}{11},
  \bibinfo{pages}{638}.
\bibitem[{Yue et~al.(2017)Yue, Zhuang, Yeh, Xie, Ma and
  Li}]{yue2017measurements}
\bibinfo{author}{Yue, Y.}, \bibinfo{author}{Zhuang, Y.}, \bibinfo{author}{Yeh,
  A.G.}, \bibinfo{author}{Xie, J.Y.}, \bibinfo{author}{Ma, C.L.},
  \bibinfo{author}{Li, Q.Q.}, \bibinfo{year}{2017}.
\newblock \bibinfo{title}{{Measurements of POI-based mixed use and their
  relationships with neighbourhood vibrancy}}.
\newblock \bibinfo{journal}{International Journal of Geographical Information
  Science} \bibinfo{volume}{31}, \bibinfo{pages}{658--675}.
\bibitem[{Zarin et~al.(2015)Zarin, Niroomand and Heidari}]{zarin2015physical}
\bibinfo{author}{Zarin, S.Z.}, \bibinfo{author}{Niroomand, M.},
  \bibinfo{author}{Heidari, A.A.}, \bibinfo{year}{2015}.
\newblock \bibinfo{title}{Physical and social aspects of vitality case study:
  Traditional street and modern street in tehran}.
\newblock \bibinfo{journal}{Procedia-Social and Behavioral Sciences}
  \bibinfo{volume}{170}, \bibinfo{pages}{659--668}.
\bibitem[{Zeng et~al.(2018)Zeng, Song, He and Shen}]{zeng2018spatially}
\bibinfo{author}{Zeng, C.}, \bibinfo{author}{Song, Y.}, \bibinfo{author}{He,
  Q.}, \bibinfo{author}{Shen, F.}, \bibinfo{year}{2018}.
\newblock \bibinfo{title}{{Spatially explicit assessment on urban vitality:
  Case studies in Chicago and Wuhan}}.
\newblock \bibinfo{journal}{Sustainable cities and society}
  \bibinfo{volume}{40}, \bibinfo{pages}{296--306}.
\bibitem[{Zhang and Seto(2011)}]{zhang2011mapping}
\bibinfo{author}{Zhang, Q.}, \bibinfo{author}{Seto, K.C.},
  \bibinfo{year}{2011}.
\newblock \bibinfo{title}{Mapping urbanization dynamics at regional and global
  scales using multi-temporal dmsp/ols nighttime light data}.
\newblock \bibinfo{journal}{Remote Sensing of Environment}
  \bibinfo{volume}{115}, \bibinfo{pages}{2320--2329}.
\bibitem[{Zhang and Cheng(2020)}]{Zhang.2020}
\bibinfo{author}{Zhang, Y.}, \bibinfo{author}{Cheng, T.}, \bibinfo{year}{2020}.
\newblock \bibinfo{title}{Graph deep learning model for network-based
  predictive hotspot mapping of sparse spatio-temporal events}.
\newblock \bibinfo{journal}{Computers, Environment and Urban Systems}
  \bibinfo{volume}{79}, \bibinfo{pages}{101403}.
\newblock \DOIprefix\doi{https://doi.org/10.1016/j.compenvurbsys.2019.101403}.
\bibitem[{Zhao et~al.(2019)Zhao, Zhou, Li, Cao, He, Yu, Li, Elvidge, Cheng and
  Zhou}]{zhao2019applications}
\bibinfo{author}{Zhao, M.}, \bibinfo{author}{Zhou, Y.}, \bibinfo{author}{Li,
  X.}, \bibinfo{author}{Cao, W.}, \bibinfo{author}{He, C.},
  \bibinfo{author}{Yu, B.}, \bibinfo{author}{Li, X.}, \bibinfo{author}{Elvidge,
  C.D.}, \bibinfo{author}{Cheng, W.}, \bibinfo{author}{Zhou, C.},
  \bibinfo{year}{2019}.
\newblock \bibinfo{title}{Applications of satellite remote sensing of nighttime
  light observations: Advances, challenges, and perspectives}.
\newblock \bibinfo{journal}{Remote Sensing} \bibinfo{volume}{11},
  \bibinfo{pages}{1971}.
\bibitem[{Zheng et~al.(2017)Zheng, Deng, Jiang, Wang, Xue, Lin, Huang, Shen, Li
  and Shahtahmassebi}]{zheng2017monitoring}
\bibinfo{author}{Zheng, Q.}, \bibinfo{author}{Deng, J.},
  \bibinfo{author}{Jiang, R.}, \bibinfo{author}{Wang, K.},
  \bibinfo{author}{Xue, X.}, \bibinfo{author}{Lin, Y.}, \bibinfo{author}{Huang,
  Z.}, \bibinfo{author}{Shen, Z.}, \bibinfo{author}{Li, J.},
  \bibinfo{author}{Shahtahmassebi, A.R.}, \bibinfo{year}{2017}.
\newblock \bibinfo{title}{{Monitoring and assessing ``ghost cities'' in
  Northeast China from the view of nighttime light remote sensing data}}.
\newblock \bibinfo{journal}{Habitat International} \bibinfo{volume}{70},
  \bibinfo{pages}{34--42}.
\bibitem[{Zhou et~al.(2019)Zhou, He, Cai, Wang and Su}]{zhou2019social}
\bibinfo{author}{Zhou, H.}, \bibinfo{author}{He, S.}, \bibinfo{author}{Cai,
  Y.}, \bibinfo{author}{Wang, M.}, \bibinfo{author}{Su, S.},
  \bibinfo{year}{2019}.
\newblock \bibinfo{title}{Social inequalities in neighborhood visual
  walkability: Using street view imagery and deep learning technologies to
  facilitate healthy city planning}.
\newblock \bibinfo{journal}{Sustainable Cities and Society}
  \bibinfo{volume}{50}, \bibinfo{pages}{101605}.

\end{thebibliography}

\end{document}